\documentclass[runningheads]{llncs}
\usepackage{graphicx}
\usepackage[table]{xcolor}
\usepackage{tikz}
\usepackage{comment}
\usepackage{amsmath,amssymb} 
\usepackage{color}

\usepackage[pagebackref,breaklinks,colorlinks,bookmarks=false,urlcolor=black,citecolor={green!60!black}]{hyperref}
\usepackage{orcidlink}

\usepackage{booktabs}
\usepackage{times}
\usepackage{epsfig}
\usepackage{multirow}
\usepackage{enumitem}
\usepackage{bbm}

\usepackage{tabularx}

\usepackage{amsthm}
\usepackage{amsfonts}
\usepackage{dsfont}
\usepackage{bm}
\usepackage{mathtools}
\usepackage{algpseudocode}
\usepackage{algorithm}

\usepackage{subcaption}
\usepackage{xspace}
\usepackage{arydshln}

\usepackage{xcolor}
\usepackage{tcolorbox}

\usepackage{breakcites}

\usepackage{multibib}
\newcites{latex}{References}

\usepackage{filecontents}

\usepackage[accsupp]{axessibility}  

\usepackage[final]{pdfpages}

\renewcommand\vec[1]{\ensuremath\boldsymbol{#1}}
\renewcommand\cdots{...}

\newcommand{\valpha}{\boldsymbol{\alpha}}

\newcommand{\mX}{\mathbf{X}}

\newcommand{\vx}{\mathbf{x}}

\newcommand{\mbrp}[1]{\mathbb{R}_{+}^{#1}}
\newcommand{\mbr}[1]{\mathbb{R}^{#1}}

\newcommand{\vbeta}{\vec{\beta}}

\newcommand{\tAnb}{\mathcal{A}}

\newcommand{\idx}[1]{\mathcal{I}_{#1}}

\newcommand{\vpsi}{\boldsymbol{\psi}}

\newcommand{\mPsi}{\vec{\Psi}}

\DeclareMathOperator*{\argmin}{arg\,min}
\DeclareMathOperator*{\argmax}{arg\,max}

\DeclareMathOperator*{\softming}{SoftMin_\gamma}

\DeclareMathOperator*{\softminsel}{SoftMinSel_\gamma}

\newcommand{\mPi}{{\boldsymbol\Pi}}

\newcommand{\vw}{\boldsymbol{w}}

\newcommand{\vsigma}{\boldsymbol{\sigma}}

\def\eg{\emph{e.g.}}

\newcommand{\cov}{\boldsymbol{\Sigma}}

\newcommand{\mM}{\boldsymbol{M}}

\newcommand{\mD}{\boldsymbol{D}}

\newcommand{\vmu}{\boldsymbol{\mu}}

\newcommand{\stkout}[1]{{\ifmmode\text{\sout{\ensuremath{#1}}}\else\sout{#1}\fi}}

\newcommand{\commentt}[1]{}

\makeatletter
\DeclareRobustCommand\onedot{\futurelet\@let@token\bmv@onedotaux}
\def\bmv@onedotaux{\ifx\@let@token.\else.\null\fi\xspace}
\def\eg{\emph{e.g}\onedot} 
\def\ie{\emph{i.e}\onedot} 
 
\def\etc{\emph{etc}\onedot} \def\vs{\emph{vs}\onedot}
\def\wrt{w.r.t\onedot}

\makeatother

\makeatletter
{\endalign\endminipage\linebreak}
\makeatother

\makeatletter
\renewcommand{\paragraph}{%
  \@startsection{paragraph}{4}%
  {\z@}{0.75ex \@plus 1ex \@minus .2ex}{-0.3em}%
  {\normalfont\normalsize\bfseries}%
}
\makeatother

\begin{document}
\pagestyle{headings}
\mainmatter
\def\ECCVSubNumber{5350}  

\title{Uncertainty-DTW for Time Series and Sequences}

\titlerunning{Uncertainty-DTW for Time Series and Sequences}
%
\author{Lei Wang$^{\star, \dagger, \S}$\orcidlink{0000-0002-8600-7099} \and
Piotr Koniusz\thanks{Equal contribution.$\;\;$ Code: {\fontsize{8}{8}\selectfont\url{https://github.com/LeiWangR/uDTW}}.$\;\;$ This work has been accepted as an oral paper at the 17\textsuperscript{th} European Conference on Computer Vision (ECCV'22).}$^{\!\,,\S,\dagger}$\orcidlink{0000-0002-6340-5289} }
\authorrunning{Wang and Koniusz}
%
\institute{$^{\dagger}$Australian National University \;
   $^\S$Data61/CSIRO\\
   $^\S$firstname.lastname@data61.csiro.au 
}
\maketitle

\thispagestyle{empty}

\begin{abstract}
 Dynamic Time Warping (DTW) is used for matching pairs of sequences and celebrated in applications such as forecasting the evolution of time series, clustering time series or even matching  sequence pairs in few-shot action recognition. 
%
The  transportation plan of DTW contains a set of paths; each path matches frames between two sequences under a varying degree of time warping, to account for varying temporal intra-class dynamics of actions. However, as DTW  is the smallest distance among all paths, it may be affected by the feature uncertainty  which varies across time steps/frames. Thus, in this paper, we propose to  model the so-called aleatoric uncertainty of a  differentiable (soft) version of DTW.
 To this end, we model the heteroscedastic aleatoric uncertainty of each path by the product of likelihoods 
from Normal distributions, each capturing variance of pair of frames. (The path distance is the sum of base distances between features of pairs of frames of the path.) The Maximum Likelihood Estimation (MLE) applied to a path yields two terms: (i) a sum of  Euclidean distances weighted by the variance inverse, and  (ii) a sum of log-variance regularization terms.  
Thus, our uncertainty-DTW  is the smallest weighted path distance among all paths, and the regularization term (penalty for the high uncertainty) is the aggregate of  log-variances along the path. The distance and the regularization term can be used in various objectives. We showcase  forecasting the evolution of time series, estimating the Fr\'echet mean of time series, and supervised/unsupervised few-shot action recognition of the articulated human 3D body joints.
\end{abstract}

\section{Introduction}
\label{sec:intro}

 Dynamic Time Warping (DTW) \cite{marco2011icml} is a  method popular in forecasting the evolution of time series, estimating the Fr\'echet mean of time series, or classifying generally understood actions. 
 The key property of DTW is its sequence matching transportation plan that allows any two sequences that are being matched to progress at different `speeds' not only in the global sense but locally in the temporal sense. As DTW is non-differentiable,  a differentiable `soft' variant of DTW, soft-DTW \cite{marco2017icml}, uses a soft-minimum function which enables backpropagation.

The role of soft-DTW is to evaluate the (relaxed) DTW distance between a pair of sequences $\mPsi\!\equiv\![\vpsi_1,\cdots,\vpsi_\tau]\!\in\!\mbr{d'\times\tau}$, $\mPsi'\!\equiv\![{\vpsi'}_1,\cdots,{\vpsi'}_{\tau'}]\!\in\!\mbr{d'\times\tau'}$ of lengths $\tau$ and $\tau'$, respectively. Under its transportation plan $\tAnb_{\tau,\tau'}$, each path  $\mPi\!\in\!\tAnb_{\tau,\tau'}$ is evaluated to ascertain the path distance, and the smallest distance is `selected' by the soft minimum: 
%
\begin{align}
&\nonumber\\[-30pt]
& d^2_{\text{DTW}}(\mPsi,\mPsi')\!=\!\softming\left(\left[\left\langle\mPi,\mD(\mPsi,\mPsi')\right\rangle\right]_{\mPi\in\tAnb_{\tau,\tau'}}\right),  
\end{align}
%
\noindent
where $\softming(\valpha)\!=\!-\gamma\!\log\sum_i\exp(-\alpha_i/\gamma)$ is the soft minimum,  $\gamma\!\geq\!0$ controls its relaxation (hard \vs soft path selection),
 and  $\mD\!\in\!\mbrp{\tau\times\tau'}\!\!\equiv\![d^2_{\text{base}}(\vpsi_m,\vpsi'_n)]_{(m,n)\in\idx{\tau}\times\idx{\tau'}}$ contains pair-wise distances between all possible pairings of frame-wise feature representations of sequences $\mPsi$ and $\mPsi'$, 
 and 
 $d^2_{\text{base}}(\cdot,\cdot)$ may be the squared Euclidean distance.

\begin{figure}[t]
\centering
\includegraphics[trim=0cm 0cm 0cm 0cm, clip=true,width=1.0\linewidth]{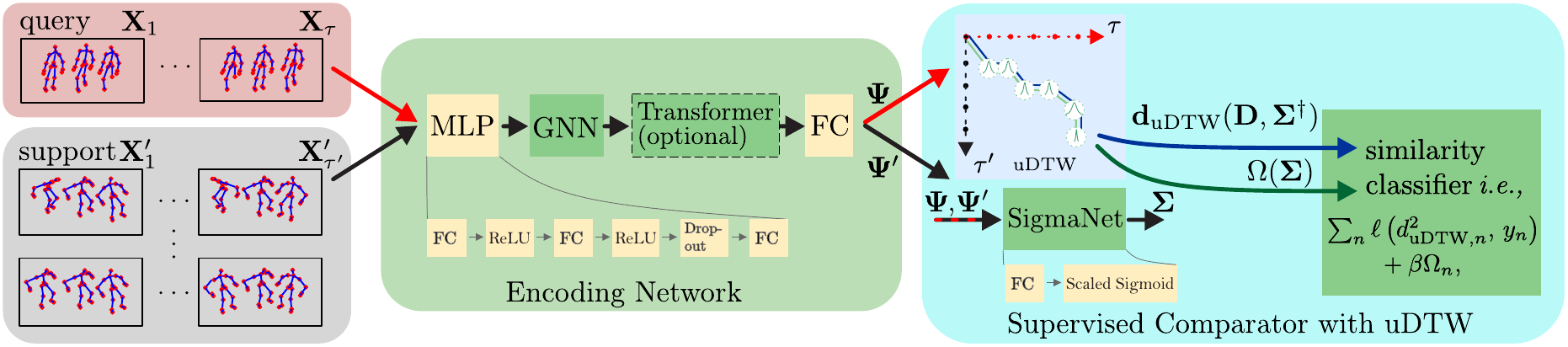}%
\caption{Supervised few-shot  action recognition of the articulated human 3D body joints with the uncertainty-DTW (uDTW).
Frames from a query and support sequences 
are split into short-term temporal blocks $\mX_1,\cdots,\mX_{\tau}$ and $\mX'_1,\cdots,\mX'_{\tau'}$ of length $M$ given stride $S$. We pass all skeleton coordinates via Encoding Network  to obtain feature tensors $\mPsi$ and $\mPsi'$, which are directed to the Supervised Comparator with uDTW. For each query-support  pair $(\mPsi_n,\mPsi'_n)$, uDTW computes the base-distance matrix $\mathbf{D}_n$ reweighted by uncertainty  $\cov^\dag_n$ to compare $\tau\!\times\!\tau'$ blocks, and SigmaNet generates underlying block-wise uncertainty parameters $\cov_n$. uDTW  finds the warping path with the smallest distance, and returns its $\Omega_n$ penalty (uncertainty aggregated along the path). 
}
\label{fig:pipe}
\end{figure}

However, the path distance $\left\langle\mPi,\mD(\mPsi,\mPsi')\right\rangle$ of path $\mPi$ ignores the observation uncertainty of  frame-wise feature representations by simply relying on the Euclidean distances stored in $\mD$. Thus, we resort to the notion of the so-called aleatoric uncertainty known from a non-exhaustive list of works about uncertainty \cite{uncertainty1,uncertainty2,uncertainty3,uncertainty4,uncertainty5}.

Specifically, to capture the aleatoric uncertainty of the Euclidean distance (or regression, \etc), one should tune the observation noise parameter of sequences. Instead of the homoscedastic model (constant observation noise), we opt for the so-called heteroscedastic aleatoric uncertainty model (the observation noise may vary with each frame/sequence). To this end, we model each path distance by the product of likelihoods of Normal distributions (we also investigate other distributions in Appendix Sec.~\ref{supp:fsl}).

\definecolor{beaublue}{rgb}{0.8, 0.85, 0.8}
\definecolor{blackish}{rgb}{0.2, 0.2, 0.2}
\vspace{0.1cm}
\begin{minipage}{0.92\linewidth}
\centering
\begin{tcolorbox}[grow to left by=0.01\linewidth, width=0.93\linewidth, colframe=blackish, colback=beaublue, boxsep=0mm, arc=2mm, left=2mm, right=2mm, top=2mm, bottom=2mm]
Our (soft) uncertainty-DTW takes the following generalized form:
\commentt{
\begin{align}
&\!\!\!\!\left\{
\begin{array}{l} 
{d^2_{\text{uDTW}}(\mD,\cov^{\dag})_{|(\mPsi,\mPsi')}}
\!=\!\softming\!\Big(\underbrace{\left[\left\langle\mPi,\mD(\mPsi,\mPsi')\!\odot\!\cov^{\dag}(\mPsi,\mPsi')\right\rangle\right]_{\mPi\in\tAnb_{\tau,\tau'}}}_{\vw}\Big)\!
\\[1pt]
\Omega({\cov})_{|(\mPsi,\mPsi')}\!=\!\softminsel\!\left(\vw, 
\left[\left\langle\mPi,\log\!\cov(\mPsi,\mPsi')\right\rangle\right]_{\mPi\in\tAnb_{\tau,\tau'}}
\right),
\end{array}
\right.\nonumber\\[-30pt]
&\label{eq:unc_dis}
\\\label{eq:unc_pen}
\end{align}
}
%
%
\begin{align}
&\!\!\!\!\left\{
\begin{array}{l} 
{d^2_{\text{uDTW}}(\mD,\cov^{\dag})}
\!=\!\softming\!\Big(\underbrace{\left[\left\langle\mPi,\mD\!\odot\!\cov^{\dag}\right\rangle\right]_{\mPi\in\tAnb_{\tau,\tau'}}}_{\vw}\Big)\!
\\[1pt]
\Omega({\cov})\!=\!\softminsel\!\left(\vw, 
\left[\left\langle\mPi,\log\!\cov\right\rangle\right]_{\mPi\in\tAnb_{\tau,\tau'}}
\right),
\end{array}
\right.\nonumber\\[-30pt]
&\label{eq:unc_dis}
\\\label{eq:unc_pen}\\
&\quad\text{where}\; \mD\!\equiv\!\mD(\mPsi,\mPsi'),\;\cov\!\equiv\!\cov(\mPsi,\mPsi')\;\,\text{and}\;\, \cov^{\dag}\!=\!\text{inv}(\cov),   \nonumber
\end{align}
\end{tcolorbox}
\end{minipage}
%

\noindent
where $\odot$ is the Hadamard product, $\cov^{\dag}(\mPsi,\mPsi')$ is the element-wise inverse of  matrix $\cov\!\in\!\mbrp{\tau\times\tau'}\!\!\equiv\![\sigma^2(\vpsi_m,\vpsi'_n)]_{(m,n)\in\idx{\tau}\times\idx{\tau'}}$ which contains pair-wise variances between all possible pairings of frame-wise feature representations from sequences $\mPsi$ and $\mPsi'$.
$\softming(\valpha)\!=\!\sum_i \alpha_i\frac{\exp(-(\alpha_i\!-\!\mu_\alpha)/\gamma)}{\sum_j\exp(-(\alpha_j\!-\!\mu_\alpha)/\gamma)}$ with $\mu_\alpha$ (the mean over  coefficients of $\valpha$) subtracted  from each coefficient $\alpha_i$ to attain stability of the softmax (into which we feed $(\alpha_i\!-\!\mu_\alpha)$). Moreover, $\softminsel(\valpha,\vbeta)\!=\!\sum_i \beta_i\frac{\exp(-(\alpha_i\!-\!\mu_\alpha)/\gamma)}{\sum_j\exp(-(\alpha_j\!-\!\mu_\alpha)/\gamma)}$ is a soft-selector returning $(\beta_{i^*}\!\!: i^*\!=\!\argmin_{i}\alpha_{i})$ if $\gamma$ approaches zero.

Eq. \eqref{eq:unc_dis} yields the uncertainty-weighted time warping distance $d^2_{\text{uDTW}}(\mD,\cov^{\dag})$ between sequences $\mPsi$ and $\mPsi'$ because $\mD$ and $\cov^{\dag}$ are both functions of $(\mPsi, \mPsi')$. 

Eq. \eqref{eq:unc_pen} provides the regularization penalty  $\Omega(\cov)$ for sequences $\mPsi$ and $\mPsi'$ (as $\cov$ is a function of $(\mPsi, \mPsi')$) which is the aggregation of log-variances along the path with the smallest distance,  \ie, path matrix  $((\mPi_{i^*}\!\in\!\{0,1\}^{\tau\times\tau'}\!)\!\!: i^*\!=\!\argmin_k w_k)$ if $\gamma\!=\!0$, and vector $\vw$ contains path-aggregated distances for all possible paths of the plan $\tAnb_{\tau,\tau'}$.

\paragraph{Contributions.}
The celebrated DTW warps the matching path between a pair of sequences to recover the best matching distance under varying temporal within-class  dynamics of each sequence. The recovered path, and  the distance corresponding to that path, may be suboptimal if frame-wise (or block-wise) features contain noise (frames that are outliers, contain occlusions or large within-class object variations,  \etc) 

\vspace{0.3cm}
To this end, we  propose several contributions:
\renewcommand{\labelenumi}{\roman{enumi}.}
\begin{enumerate}[leftmargin=0.6cm]
\item  We introduce the uncertainty-DTW, dubbed as uDTW, whose role is to take into account the uncertainty of in frame-wise (or block-wise) features by selecting the path which maximizes the Maximum Likelihood Estimation (MLE). The parameters (such as variance) of a distribution (\ie, the Normal distribution) are thus used within MLE (and uDTW) to model the uncertainty.
\item As pairs of sequences are often of different lengths, optimizing the free-form variable of variance is impossible. To that end, we equip each of our pipelines with SigmaNet, whose role is to take frames (or blocks) of sequences, and generate the variance end-to-end (the variance is parametrized by SigmaNet).
\item We provide several pipelines that utilize uDTW for (1) forecasting the evolution of time series, (2) estimating the Fr\'echet mean of time series, (3) supervised few-shot action recognition, and (4) unsupervised  few-shot action recognition.
\end{enumerate}

\paragraph{Notations.} 
$\idx{\tau}$ is the index set $\{1,2,\cdots,\tau\}$. Concatenation of $\alpha_i$ into a vector $\valpha$ is denoted by $[\alpha_i]_{i\in\idx{I}}$. Concatenation of $\alpha_{ij}$ into  matrix $\mathbf{A}$ is denoted by $[\alpha_{ij}]_{(i,j)\in\idx{I}\times\idx{J}}$. Dot-product between two matrices  equals the dot-product of vectorized $\mPi$ and $\mD$, that is  $\left\langle\mPi,\mD\right\rangle\!\equiv\!\left\langle\text{vec}({\mPi}),\text{vec}(\mD)\right\rangle$.  
%
Mathcal symbols are sets, \eg, $\tAnb$ is a transportation plan, capitalized bold symbols are matrices, \eg, $\mD$ is the distance matrix, lowercase bold symbols  are vectors, \eg, $\vw$ contains weighted distances. Regular fonts are scalars.

\begin{figure*}[t]
\centering
%
\begin{subfigure}[t]{0.195\linewidth}
\centering\includegraphics[trim=2.3cm 0.258cm 2.3cm 0.258cm, clip=true, width=\linewidth,height=1.8cm]{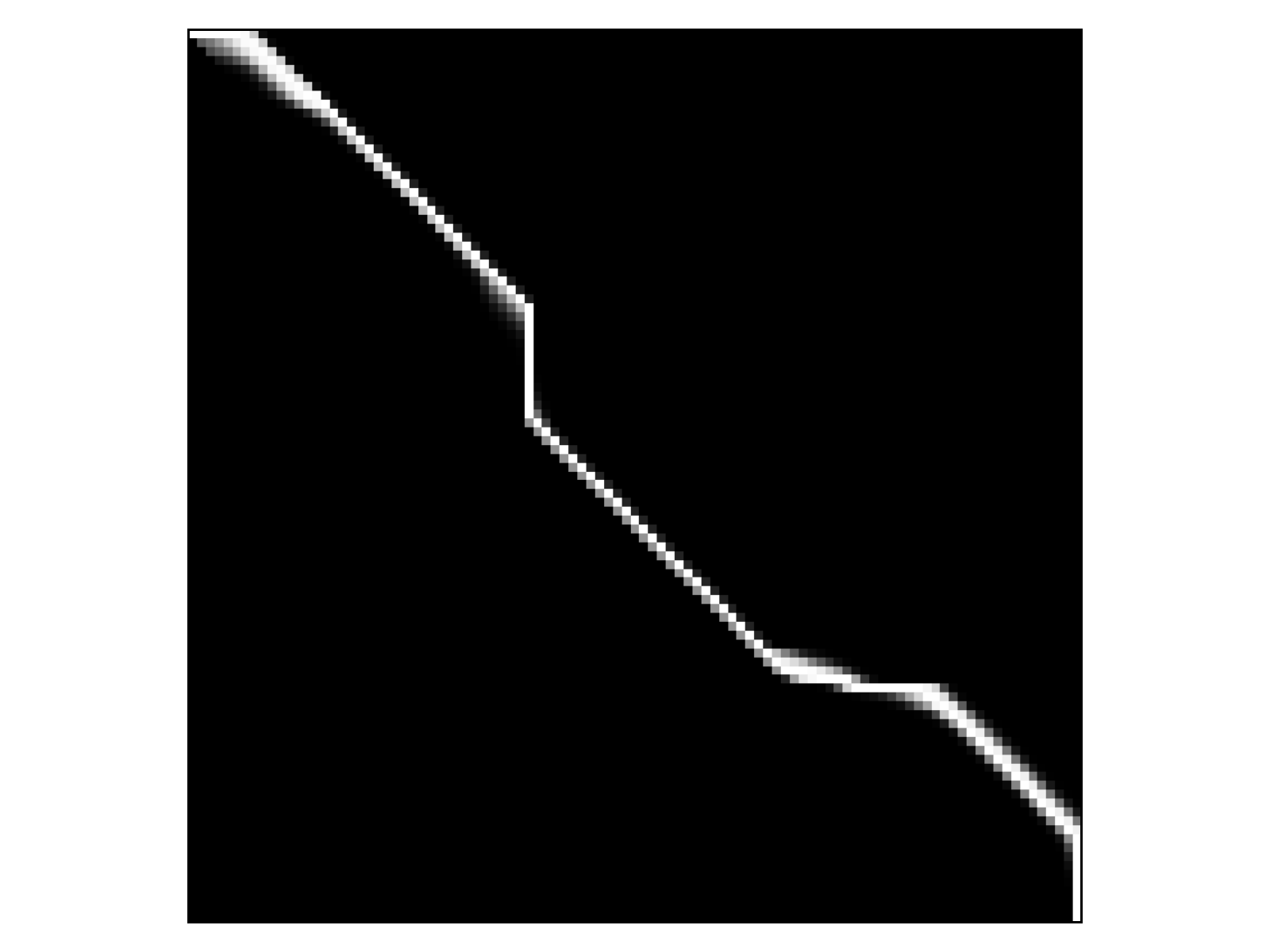}
\caption{$\text{sDTW}_{\gamma=0.01}$}\label{fig:sdtw_0.01}
\end{subfigure}
\begin{subfigure}[t]{0.195\linewidth}
\centering\includegraphics[trim=2.3cm 0.258cm 2.3cm 0.258cm,clip=true, width=\linewidth,height=1.8cm]{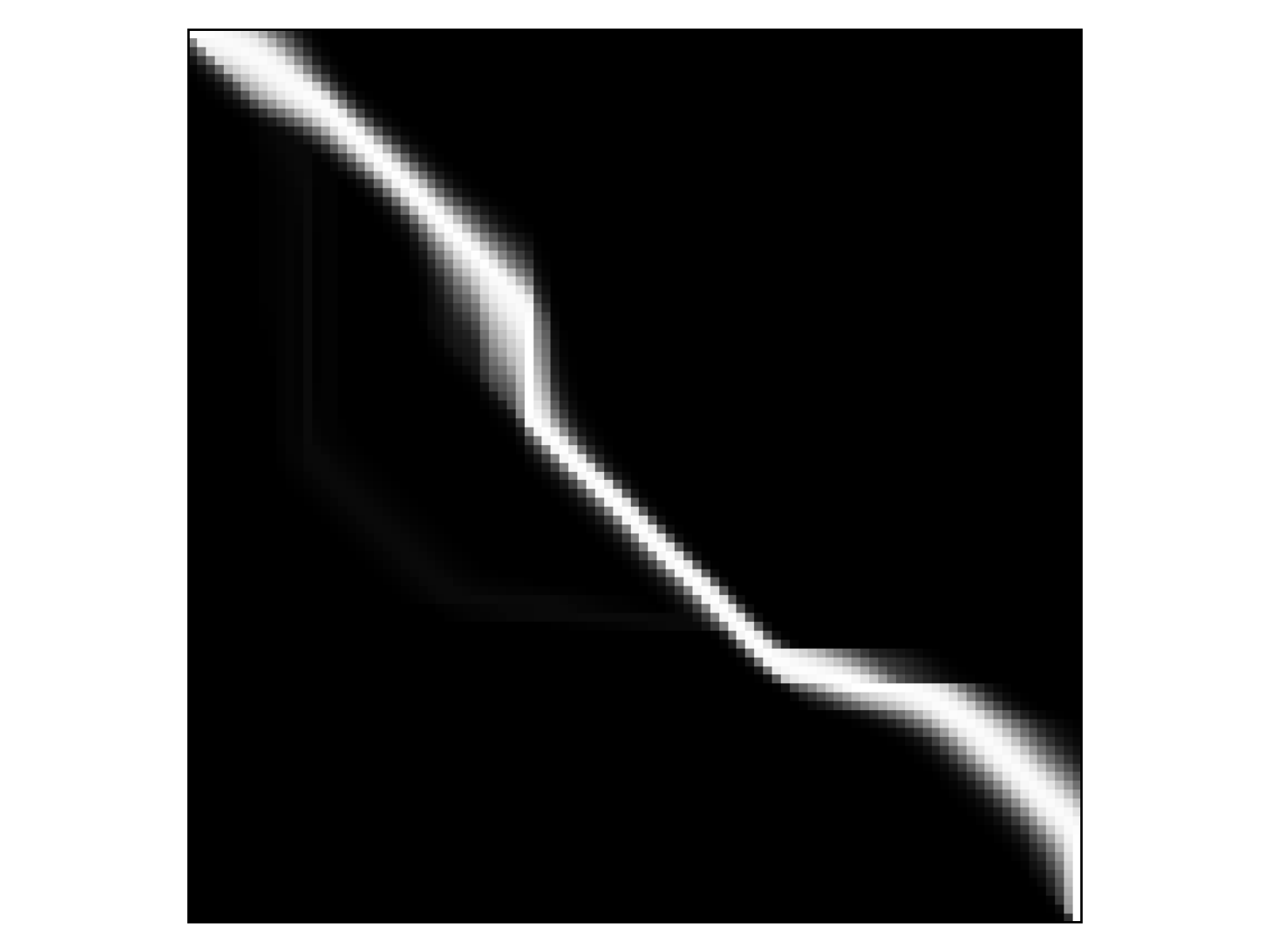}
\caption{$\text{sDTW}_{\gamma=0.1}$}\label{fig:sdtw_0.1}
\end{subfigure}
\begin{subfigure}[t]{0.195\linewidth}
\centering\includegraphics[trim=2.3cm 0.258cm 2.3cm 0.258cm,clip=true, width=\linewidth,height=1.8cm]{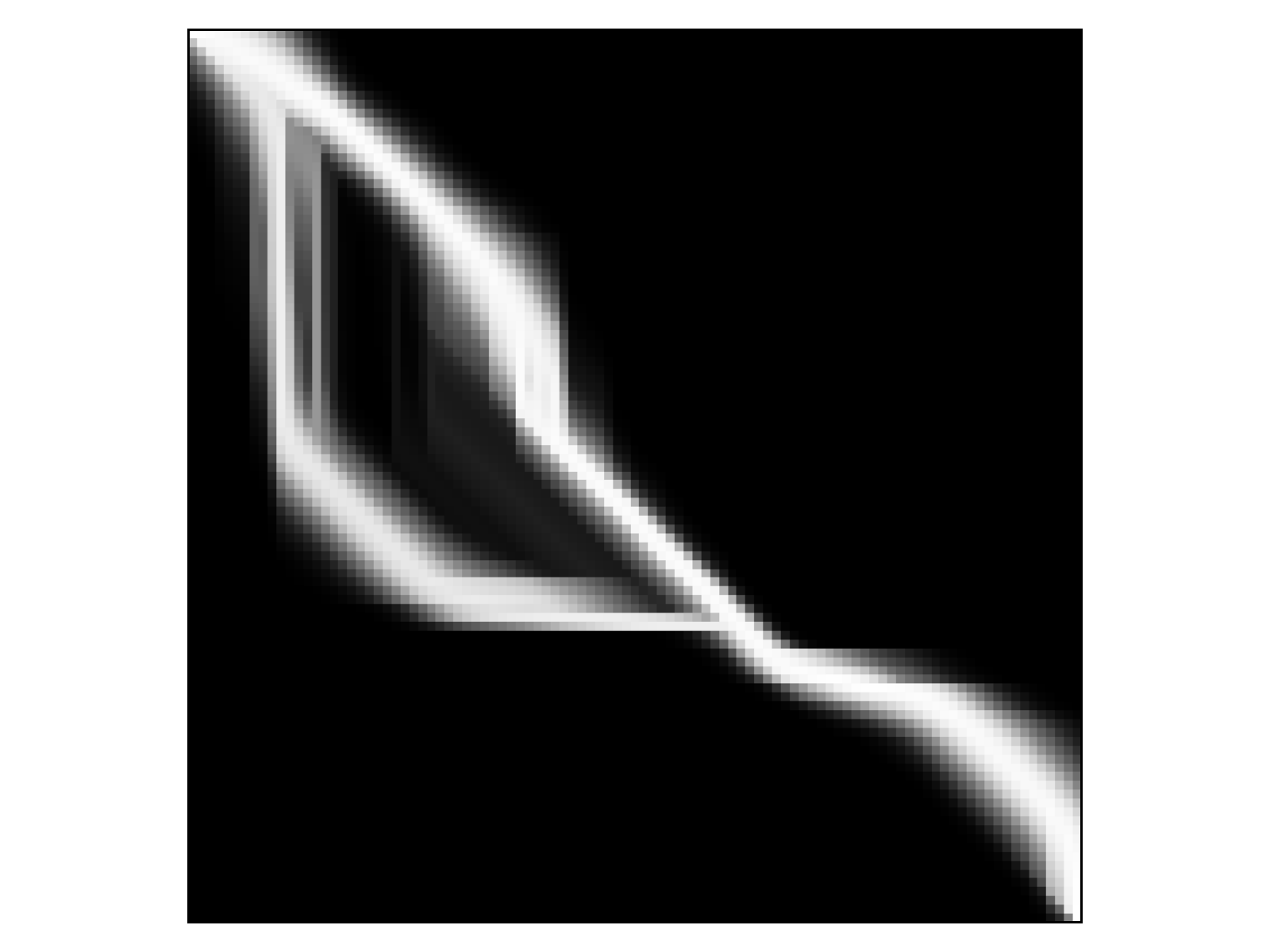}
\caption{$\text{uDTW}_{\gamma=0.01}$}\label{fig:udtw_0.01}
\end{subfigure}
\begin{subfigure}[t]{0.195\linewidth}
\centering\includegraphics[trim=2.3cm 0.255cm 2.3cm 0.255cm, clip=true, width=\linewidth,height=1.8cm]{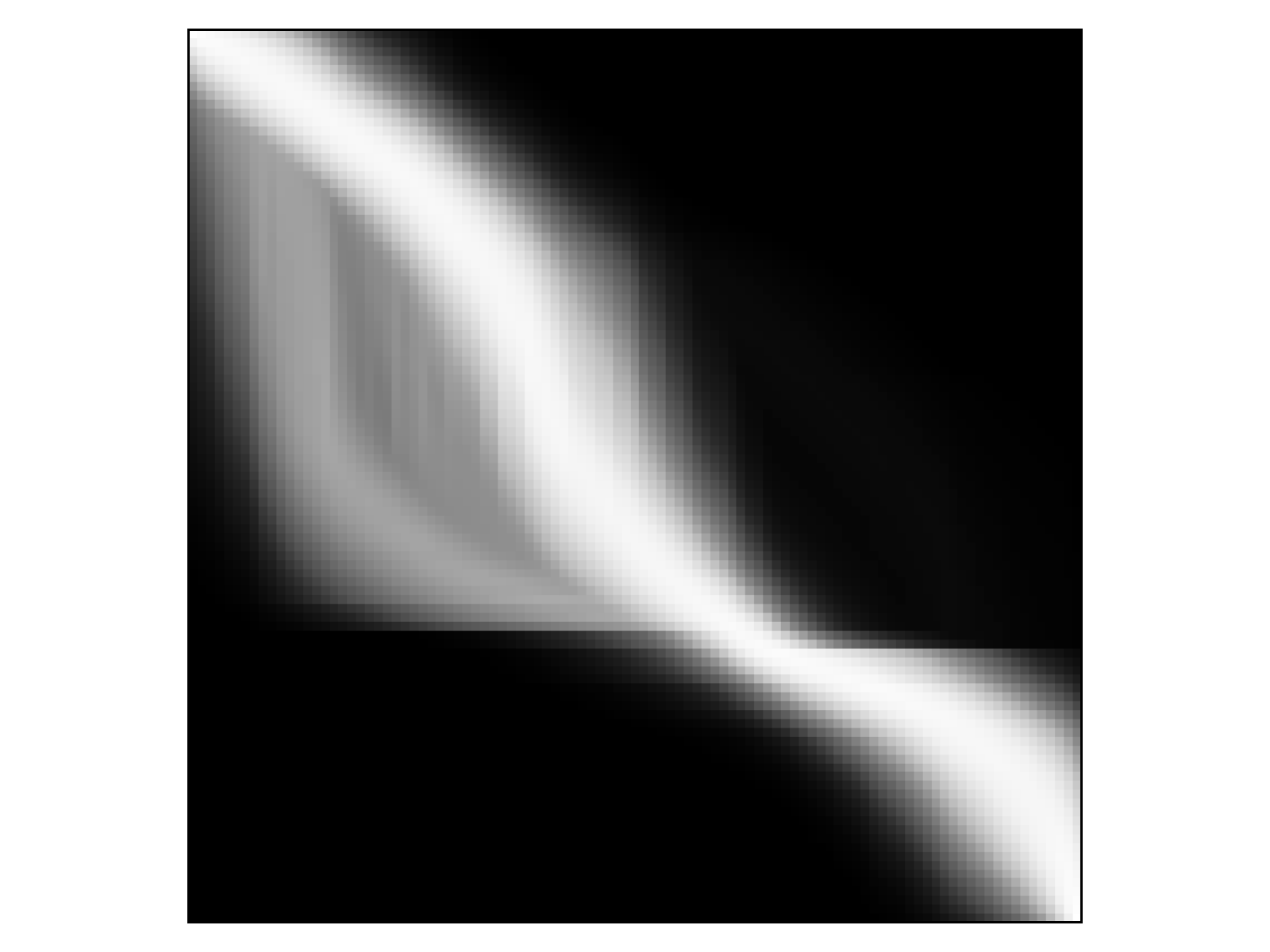}
\caption{$\text{uDTW}_{\gamma=0.1}$}\label{fig:udtw_0.1}
\end{subfigure}
\begin{subfigure}[t]{0.195\linewidth}
\centering\includegraphics[trim=8cm 3.48cm 8cm 3.88cm, clip=true, width=\linewidth,height=1.8cm]{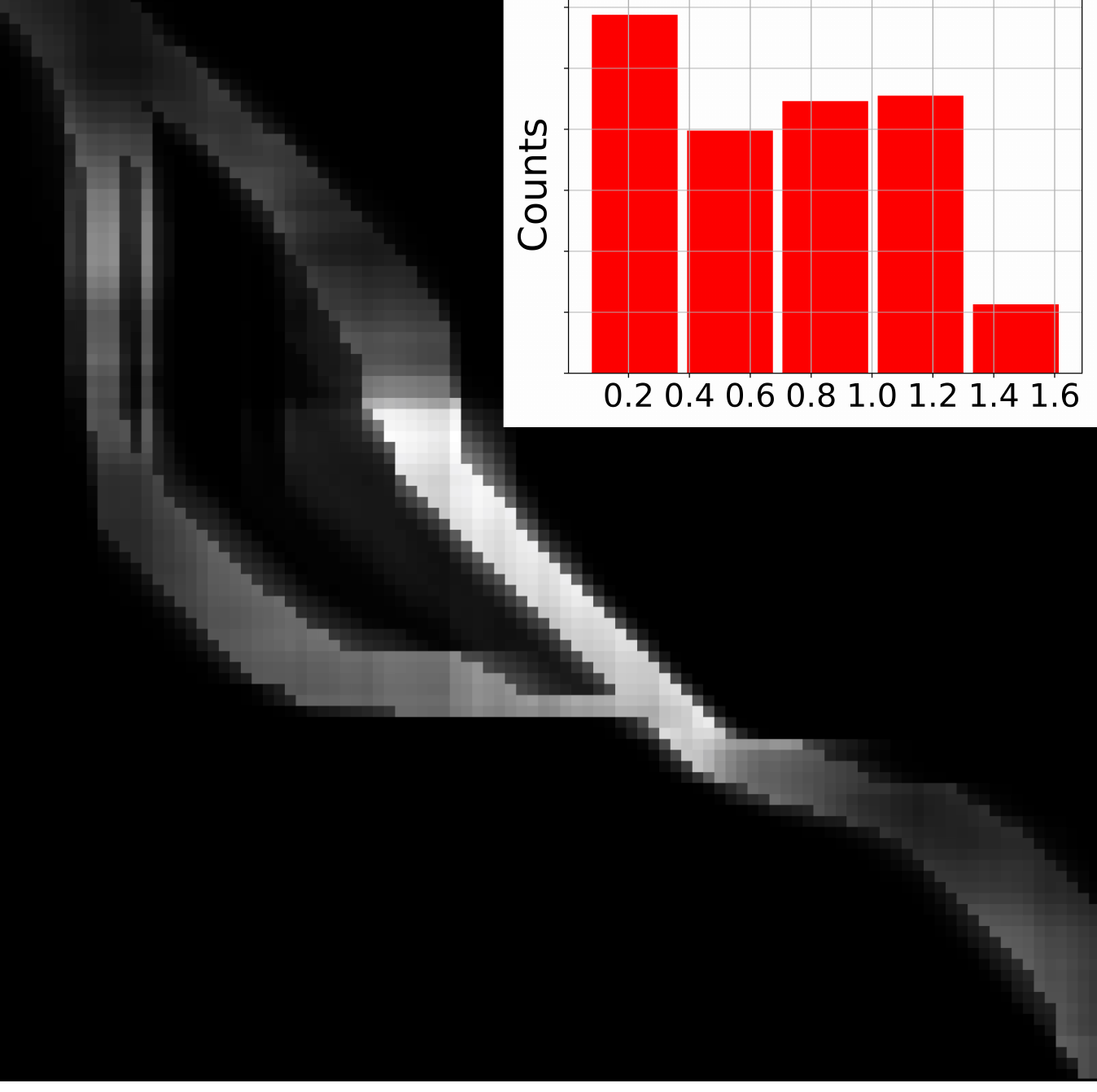}
\caption{uDTW uncert.}\label{fig:udtw_cov_hiscounts}
\end{subfigure}
%
%
%
%
\caption{Plots (a)-(d) show  paths of sDTW and uDTW (in white) for a  pair of sequences. We power-normalized pixels of plots (by the power of 0.1) to see also darker paths better.  
With higher $\gamma$ that controls softness, in (b) \& (d) more paths  become `active' (fuzzy effect). 
In (c), uDTW has two possible routes \vs sDTW (a) due to uncertainty modeling. 
In (e), we visualise uncertainty $\cov$. We binarize plot (c) and multiply it by the $\cov$ to display uncertainty values on the path (white pixels = high uncertainty). The middle of the main path is deemed uncertain, which explains why an additional path merges in that region with the main path. See also the histogram of values of $\cov$.
%
}
\vspace{-0.1cm}
\label{fig:visual}
\end{figure*}

\subsection{Similarity learning with uDTW}
In further chapters, based on the distance in Eq. \eqref{eq:unc_dis} and the regularization term in Eq.  \eqref{eq:unc_pen}, we define specific loss functions for several problems  such as  forecasting the evolution of time series, clustering time series or even matching sequence pairs in few-shot action recognition. Below is an example of a generic similarity learning loss:
\begin{align}
&\label{eq:loss1}\argmin_{\mathcal{P}} \sum_{n} \ell\left(d^2_{\text{uDTW}}(\mD(\mPsi_n,\mPsi'_{n}), \cov^{\dag}(\mPsi_n,\mPsi'_{n}) ),\, \delta_n\right)+\beta\Omega(\cov(\mPsi_n,\mPsi'_n)),\\[-7pt]
&\text{or}\nonumber\\[-3pt]
&\label{eq:loss2}\argmin_{\mathcal{P},\cov>0} \sum_{n} \ell\left(d^2_{\text{uDTW}}(\mD(\mPsi_n,\mPsi'_{n}), \cov^{\dag} ),\, \delta_n\right)+\beta\Omega(\cov),
\end{align}
where $\mPsi_n\!=\!f(\mX_n;\mathcal{P})$ and $\mPsi'_n\!=\!f(\mX'_n;\mathcal{P})$ are obtained from some backbone encoder $f(\cdot;\mathcal{P})$ with parameters $\mathcal{P}$ and $(\mX_n,\mX'_n)\!\in\!\mathcal{X}$ is a sequence pair to compare with the similarity label $\delta_n\!\in\!\{0,1\}$ (where $\delta_n\!=\!0$ if $y_n\!=\!y'_n$ and $\delta_n\!=\!1$ otherwise), $(y_n,y'_n)$ is a pair of class labels for $(\mPsi_n,\mPsi'_n)$, and $\beta\!\geq\!0$ controls the penalty for high matching uncertainty. Figure \ref{fig:visual} illustrates the impact of uncertainty on uDTW.

{
\vspace{0.1cm}
\begin{minipage}{0.92\linewidth}
\centering
\begin{tcolorbox}[grow to left by=0.01\linewidth, width=1\linewidth, colframe=blackish, colback=beaublue, boxsep=0mm, arc=2mm, left=2mm, right=2mm, top=2mm, bottom=2mm]
Note that minimizing Eq. \eqref{eq:loss2} \wrt $(\mathcal{P}, \cov)$ 
assumes that $\cov\in\mbrp{\tau\times\tau'}$ is a free  variable to minimize over (derivation in Section \ref{sec:der}). However, as sequence pairs vary in length, \ie, $\tau\!\neq\!\tau'$, optimizing one global $\cov$  is impossible (its size changes). Thus, for  problems we tackle, we minimize loss functions with the distance/penalty in Eq. \eqref{eq:loss1} and \eqref{eq:loss2}  where $\cov$ is parametrized by $(\mPsi_n,\mPsi'_{n})$:
%
%
\begin{align}
&\nonumber\\[-30pt]
    &d^2_{\text{uDTW}^\bullet}(\mPsi,\mPsi')\!\equiv\!d^2_{\text{uDTW}}(\mD(\mPsi,\mPsi'),\cov^{\dag}(\mPsi,\mPsi')),\\ 
 &\Omega_\bullet(\mPsi,\mPsi')\!\equiv\!\Omega(\cov(\mPsi,\mPsi')).
\end{align}

To that end, we devise a small MLP unit  $\sigma(\cdot; \mathcal{P}_\sigma)$ or $\sigma(\cdot,\cdot; \mathcal{P}_\sigma)$ and obtain:$\!\!\!\!$
\begin{align}
    & \label{eq:add}\cov\!=\!0.5\!\cdot\![(\sigma^2(\vpsi_m; \mathcal{P}_\sigma)+\sigma^2(\vpsi'_n;\mathcal{P}_\sigma))]_{(m,n)\in\idx{\tau}\times\idx{\tau'}}\\[-5pt]
&\text{or}\nonumber\\[-3pt]
&\label{eq:joint}\cov'\!=\![\sigma^2(\vpsi_m,\vpsi'_n; \mathcal{P}_\sigma)]_{(m,n)\in\idx{\tau}\times\idx{\tau'}},
\end{align}
where Eq. \eqref{eq:add} uses additive variance terms generated for individual frames $\vpsi_m$ and $\vpsi'_n$, whereas \eqref{eq:joint} is a jointly generated variance for $(\vpsi_m, \vpsi'_n)$.
\end{tcolorbox}
\end{minipage}
\vspace{0.1cm}
}


\subsection{Derivation of uDTW}
\label{sec:der}

We proceed by modeling an arbitrary path $\mPi_i$ from the transportation plan of $\tAnb_{\tau,\tau'}$ as the following Maximum Likelihood Estimation (MLE) problem:
\begin{align}
& \argmax\limits_{\{\sigma_{mn}\}_{(m,n)\in\mPi_i}} \prod_{(m,n)\in\mPi_i}p(\|\vpsi_m\!-\!\vpsi'_n\|, \sigma^2_{mn}),
\end{align}
where $p$ may be some arbitrary distribution, $\sigma$ are distribution parameters,  and $\|\cdot\|$ is an arbitrary norm. For the Normal distribution $\mathcal{N}$ which relies on the squared Euclidean distance $\|\cdot\|_2^2$, we have:
\begin{align}
& \argmax\limits_{\{\sigma_{mn}\}_{(m,n)\in\mPi_i}} \prod_{(m,n)\in\mPi_i}\mathcal{N}(\vpsi_m; \vpsi'_n, \sigma^2_{mn})\\
&\quad=\argmax\limits_{\{\sigma_{mn}\}_{(m,n)\in\mPi_i}} \log\prod_{(m,n)\in\mPi_i}\frac{1}{(2\pi)^{\frac{d'}{2}}\sigma^{d'}}\exp\Big(\!-\frac{\|\vpsi_m\!-\!\vpsi'_n\|_2^2}{\sigma^2_{mn}}\Big)\\
&\quad=\argmax\limits_{\{\sigma_{mn}\}_{(m,n)\in\mPi_i}} \sum_{(m,n)\in\mPi_i} -\frac{d'}{2}\log(2\pi) -d'\log(\sigma) -\frac{\|\vpsi_m\!-\!\vpsi'_n\|_2^2}{\sigma^2_{mn}}\\
&\quad=\argmin\limits_{\{\sigma_{mn}\}_{(m,n)\in\mPi_i}} \sum_{(m,n)\in\mPi_i} d'\log(\sigma) +\frac{\|\vpsi_m\!-\!\vpsi'_n\|_2^2}{\sigma^2_{mn}},
\end{align}
where $d'$ is the length of feature vectors $\vpsi$. Having recovered uncertainty parameters ${\{\sigma_{mn}\}_{(m,n)\in\mPi_i}}$, we obtain a combination of penalty terms and reweighted squared Euclidean distances:
\begin{align}
& \beta\Omega_{\mPi_i} +d^2_{\mPi_i}\!=\!\sum_{(m,n)\in\mPi_i} \beta\log(\sigma_{mn}) +\frac{\|\vpsi_m\!-\!\vpsi'_n\|_2^2}{\sigma^2_{mn}},
\end{align}
where $\beta\!\geq\!0$ (generally $\beta\!\neq\!d'$) adjusts the penalty for large uncertainty. Separating the uncertainty penalty $\log(\sigma_{mn})$ from the uncertainty-weighted distance (both aggregated along path $\mPi_i$) yields:
\begin{align}
\left\{
\begin{array}{l} 
d^2_{\mPi_i}\!=\!\left\langle\mPi_i,\mD(\mPsi,\mPsi')\!\odot\!\cov^{\dag}
\right\rangle\\
\Omega_{\mPi_i}\!=\!\left\langle\mPi_i,\log\!\cov\right\rangle,
\end{array}
\right.
\end{align}
where $\mD\!\in\!\mbrp{\tau\times\tau'}\!\!\equiv\!\big[\frac{d^2_{2}(\vpsi_m,\vpsi'_n)}{\sigma^2_{mn}}\big]_{(m,n)\in\idx{\tau}\times\idx{\tau'}}$ and $\cov\!\in\!\mbrp{\tau\times\tau'}\!\!\equiv\![\sigma^2_{mn}]_{(m,n)\in\idx{\tau}\times\idx{\tau'}}$. Derivations for other distributions, \ie, Laplace or Cauchy, follow the same reasoning.

\section{Related Work}
\label{sec:related}

\noindent{\bf Different flavors of Dynamic Time Warping.} DTW~\cite{marco2011icml}, which seeks a minimum cost alignment between time series is computed by dynamic programming in quadratic time, is not differentiable and is known to get trapped in bad local minima. 
In contrast, soft-DTW (sDTW)~\cite{marco2017icml} addresses the above issues by replacing the minimum over alignments with a soft minimum, which has the effect of inducing a `likelihood' field over all possible alignments. However, sDTW has been successfully applied in many computer vision tasks including audio/music score alignment~\cite{pmlr-v80-mensch18a}, action recognition~\cite{su2022temporal,kaidi2020cvpr}, and end-to-end differentiable text-to-speech synthesis~\cite{donahue2021endtoend}. 
Despite its  successes, sDTW has some limitations: (i) it can be negative when used as a loss (ii) it may still get trapped in bad local minima. 
Thus, soft-DTW divergences (sDTW div.)~\cite{pmlr-v130-blondel21a}, inspired by sDTW, attempts to overcome such issues. 

Other  approaches inspired by DTW have been used to improve  the inference or adapt to modified or additional constraints,  \ie, 
OPT~\cite{8466003} and OWDA~\cite{9008561} treat the alignment as the optimal transport problem with temporal regularization. TAP~\cite{su2022temporal} directly predicts the alignment through a lightweight CNN, thus is does not follow a principled transportation plan, and is not guaranteed to find a minimum cost path.

Our uDTW differs from these methods in that the transportation plan is executed under the uncertainty estimation, thus various feature-level noises and outliers are less likely to lead to the selection of a sub-optimal cost path.

\noindent{\bf Alignment-based time series problems.} Distance between sequences plays an important role in time series retrieval~\cite{9008561}, forecasting~\cite{marco2017icml,pmlr-v130-blondel21a}, classification~\cite{marco2017icml,pmlr-v130-blondel21a,10.1145/3447548.3467231,pmlr-v139-yang21j}, clustering~\cite{4668349,1163055}, \etc. Various temporal nuisance noises such as initial states, different sampling rates, local distortions, and execution speeds make the measurement of distance between sequences difficult. To tackle these issues, typical feature-based methods use RNNs to encode sequences and measure the distance between corresponding features~\cite{ramachandran2018unsupervised}.     Other existing methods~\cite{lei_thesis_2017,lei_icip_2019,hosvd} either encode each sequence into features that are invariant to temporal variations~\cite{10.5555/3327546.3327715,Lohit_2019_CVPR} or adopt alignment for temporal correspondence calibration \cite{8466003}. However, none of these methods is modeling the aleatoric uncertainty. As we model it along the time warping path, the observation noise may vary with each frame or block.

\noindent{\bf Few-shot action recognition.} Most existing few-shot action recognition methods~\cite{lei_tip_2019,lei_iccv_2019,10.1145/3474085.3475572} follow the metric learning  paradigm. Signal Level Deep Metric Learning \cite{2021dml}  and Skeleton-DML \cite{memmesheimer2021skeletondml} one-shot FSL approaches  encode  signals  into  images,  extract  features  using  a  deep  residual  CNN and apply multi-similarity miner losses. TAEN~\cite{9523103} and FAN~\cite{8851694} encode actions into representations and apply vector-wise metrics. 

Most methods identify the importance of temporal alignment for handling the non-linear temporal variations, and various alignment-based models are proposed to compare the sequence pairs, \eg, permutation-invariant spatial-temporal attention reweighted distance in ARN~\cite{hongguang2020eccv}, a variant of DTW used in OTAM~\cite{kaidi2020cvpr}, temporal attentive relation network~\cite{tarn_BMVC2019}, a two-stage temporal alignment network (TA2N)~\cite{ta2n2021}, a temporal CrossTransformer~\cite{Perrett_2021_CVPR}, a learnable sequence matching distance called TAP~\cite{su2022temporal}. 

In all cases, temporal alignment is a well-recognized tool, however lacking the uncertainty modeling, which impacts the quality of alignment. Such a gap in the literature inspires our work on uncertainty-DTW.

\begin{figure*}[t]
\centering
%
\begin{subfigure}[t]{0.49\linewidth}
\centering\includegraphics[trim=0cm 0cm 0cm 0cm, clip=true,height=2.5cm]{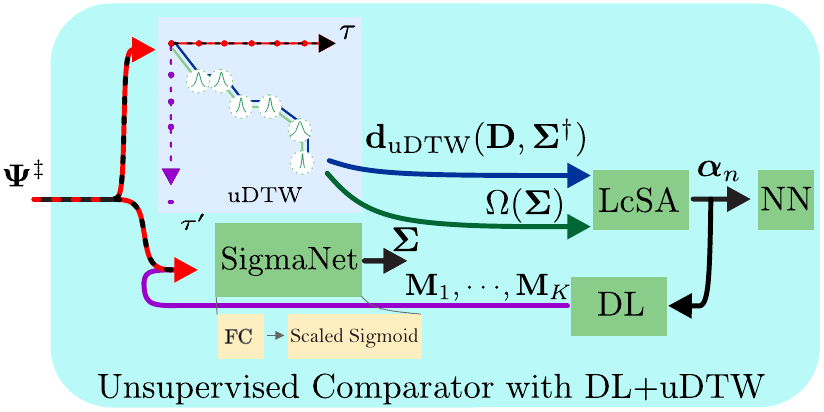}
\caption{\label{fig:unsup}}
\end{subfigure}
\begin{subfigure}[t]{0.49\linewidth}
\centering\includegraphics[trim=0cm 0cm 0cm 0cm,clip=true,height=2.5cm]{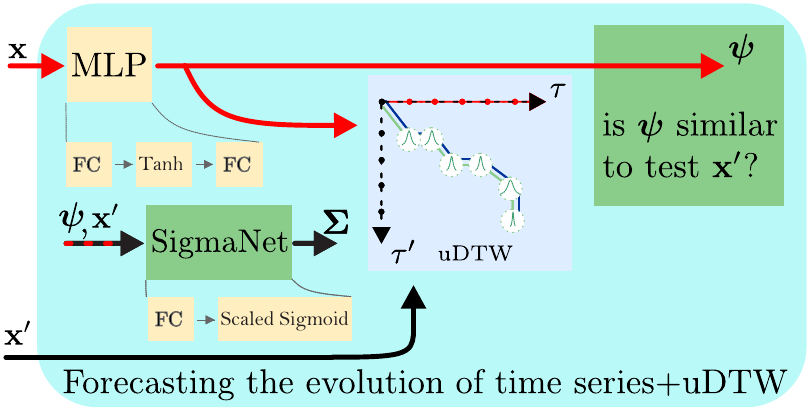}
\caption{\label{fig:forecast}}
\end{subfigure}
\caption{In (a) is the unsupervised comparator for unsupervised few-shot action recognition. The unsupervised head is wired with the Encoding Network from Figure \ref{fig:pipe}, and trained from scratch without labels. In (b) is the pipeline for forecasting the evolution of time series (a.k.a. multistep-ahead prediction).}
\label{fig:unsup_forecast}
\end{figure*}

\section{Pipeline Formulations}
\label{sec:approach}

Below we provide our several pipeline formulations for which uDTW is used as an indispensable component embedded with the goal of measuring the distance for warped paths under uncertainty.

\subsection{Few-shot Action Recognition}
\label{sec:fsl}

For both supervised and unsupervised few-shot pipelines, we employ the Encoder Network (EN) and the Supervised Comparator (similarity learning) as in Figure \ref{fig:pipe}, or  Unupervised Comparator (based on dictionary learning) as in Figure \ref{fig:unsup}.  

\paragraph{Encoding Network (EN).}  Our EN contains a simple 3-layer MLP unit (FC, ReLU, FC, ReLU, Dropout, FC), GNN, with  transformer~\cite{dosovitskiy2020image} 
and FC. 
The MLP unit takes $M$ neighboring frames, each with $J$ skeleton body joints given by Cartesian coordinates $(x,y,z)$, forming one temporal block\footnote{We use temporal blocks as they were shown more robust than frame-wise  FSAR \cite{hongguang2020eccv} models.}. In total, depending on stride $S$, we obtain some $\tau$ temporal blocks (each block captures the short temporal dependency), whereas the long temporal dependency will be modeled by uDTW. Each temporal block is encoded by the MLP into a $d\!\times\!J$ dimensional feature map.  Subsequently,  query feature maps of size $\tau$ and support feature maps of size $\tau'$  are  forwarded to a simple linear GNN model, and transformer, 
and  an FC layer, which returns $\mPsi\!\in\!\mbr{d'\times\tau}$ query feature maps and $\mPsi'\!\in\!\mbr{d'\times\tau'}\!$ support feature maps.
Such encoded feature maps are  passed to the Supervised Comparator with uDTW. 

Let support  maps  $\mPsi'\!\equiv\![f(\boldsymbol{X}'_1;\mathcal{P}),\cdots,f(\boldsymbol{X}'_{\tau'};\mathcal{P})]\!\in\!\mbr{d'\times\tau'}\!$
and query maps $\mPsi\!\equiv\![f(\boldsymbol{X}_1;\mathcal{P}),\cdots,f(\boldsymbol{X}_\tau;\mathcal{P})]\!\in\!\mbr{d'\times\tau}$ for  
query and support frames per block $\mX,\mX'\!\in\!\mbr{3\times J\times M}$. Define {\fontsize{9}{9}\selectfont$f(\mX; \mathcal{P})\!=\!\text{FC}(\text{Transf}(\text{S\textsuperscript{2}GC}(\text{MLP}(\mX; \mathcal{P}_{MLP}); \mathcal{P}_{S^2GC});$
$\mathcal{P}_{Transf}); \mathcal{P}_{FC})$}
where $\mathcal{P}\!\equiv\![\mathcal{P}_{MLP},\mathcal{P}_{S^2GC},\mathcal{P}_{Transf},\mathcal{P}_{FC},\mathcal{P}_{SN}]$ is the  set of  parameters of EN, where $\mathcal{P}_{SN}$ are parameters of  SigmaNet, 
and S\textsuperscript{2}GC is a Simple Spectral Graph Convolution (S$^2$GC) \cite{hao2021iclr} whose details are in Sec. \ref{sec:ssgc} of the Appendix.
%

\paragraph{Supervised Few-shot Action Recognition.} For the $N$-way $Z$-shot problem, 
we have   one query feature map  and $N\!\times\!Z$ support feature maps per episode. We form a mini-batch containing $B$ episodes. We have query feature maps $\{\mPsi_b\}_{b\in\idx{B}}$ and support feature maps $\{\mPsi'_{b,n,z}\}_{b\in\idx{B},n\in\idx{N},z\in\idx{Z}}$. Moreover,  $\mPsi_b$ and $\mPsi'_{b,1,:}$ share the same class (drawn from $N$ classes per episode), forming the  subset $C^{\ddagger} \equiv \{c_1,\cdots,c_N \} \subset \mathcal{I}_C \equiv \mathcal{C}$. To be precise, labels  $y(\mPsi_b)\!=\!y(\mPsi'_{b,1,z}), \forall b\!\in\!\idx{B}, z\!\in\!\idx{Z}$ while $y(\mPsi_b)\!\neq\!y(\mPsi'_{b,n,z}), \forall b\!\in\!\idx{B},n\!\in\!\idx{N}\!\setminus\!\{1\},  z\!\in\!\idx{Z}$. Thus the similarity label $\delta_1\!=\!0$, whereas $\delta_{n\neq1}\!=1\!$. 
%
%
Note that the selection of $C^{\ddagger}$ per episode is random. For the $N$-way $Z$-shot protocol, the Supervised Comparator is minimized \wrt $\mathcal{P}$ ($\mPsi_b$ and $\mPsi'$ depend on $\mathcal{P}$) as:
%
\begin{align}
    &\argmin_{\mathcal{P}}\sum_{b\in\idx{B}}\sum_{n\in\idx{N}}\sum_{z\in\idx{Z}}\left(d^2_{\text{uDTW}^\bullet}(\mPsi_b,\mPsi'_{b,n,z})-\delta_n\right)^2\!+\beta\Omega_\bullet(\mPsi_b,\mPsi'_{b,n,z}).
    \label{eq:sup_ar}
\end{align}

\paragraph{Unsupervised Few-shot Action Recognition.} Below we propose a very simple unsupervised variant with so-called Unsupervised Comparator.  The key idea is that with uDTW, invariant to local temporal speed changes can be used to learn a dictionary which, with some dictionary coding method should outperform at reconstructing the sequences. This means we can learn an unsupervised comparator by projecting sequences onto the dictionary space. To this end, let the protocol remain as for the supervised few-shot learning with the exception that class labels are not used during training, and only support images in testing are labeled for sake of evaluation the accuracy by deciding which support representation each query is the closest to in the nearest neighbor sense.

Firstly, in each training episode, we combine the query sequences $\mPsi_b$ with the support sequences $\mPsi'_{b,n,z}$ into episode sequences denoted as $\mPsi^\ddag_{b,n}$  where $b\!\in\!\idx{B}$ enumerates over $B$ episodes, and $n\!\in\!\idx{(N\cdot Z+1)}$. For the feature coding, we use Locality-constrained Soft Assignment (LCSA) \cite{6126534,6116129,KONIUSZ2013479} and a simple dictionary update based on the least squares computation. 

\vspace{0.2cm}
For each episode $b\!\in\!\idx{B}$, we iterate over the following three  steps:

\renewcommand{\labelenumi}{\roman{enumi}.}
\begin{enumerate}[leftmargin=0.6cm]
\item The LCSA coding step which expresses each $\mPsi^\ddag_{b,n}$ as $\valpha_{b,n}\!\in\!\mbrp{K}$ that assign  $\mPsi^\ddag_{b,n}$ into a dictionary with $K$ sequences $\mM_1,\cdots,\mM_K\!\in\!\mbr{d'\times\tau'}$ (dictionary anchors): 
\begin{align}
& \!\!\!\!\forall_{k,n},\;\alpha_{k,b,n}\!=\!\left\{\begin{array}{l}
\frac{\exp\big(-\frac{1}{\gamma'}d^2_{\text{uDTW}^\bullet}\big(\mPsi^\ddag_{b,n}, \mM_k\big)\big)}{\!\!\!\!\sum\limits_{l\in\mathcal{M}(\mPsi^\ddag_{b,n};K')}\!\!\!\!\exp\big(-\frac{1}{\gamma'}d^2_{\text{uDTW}^\bullet}\big(\mPsi^\ddag_{b,n}, \mM_l\big)\big)} \quad\text{if}\;\;\mM_k\!\in\!\mathcal{M}\big(\mPsi^\ddag_{b,n};K'\big),\\
0 \qquad\qquad\qquad\qquad\qquad\text{otherwise},
\end{array}\right.
    \nonumber\\[-16pt]
    &\label{eq:lcsa} 
\end{align}
where $0\!<\!K'\!\leq\!K$ is a subset size for $K'$ nearest anchors of $\mPsi^\ddag_{b,n}$ retrieved by operation $\mathcal{M}(\mPsi^\ddag_{b,n};K')$ (based on uDTW) from $\mM_1,\cdots,\mM_K$,  $\tau'$ is set to the mean of $\tau$ (over training set), and $\gamma'\!=\!0.7$ is a so-called smoothing factor;

\item The dictionary update step updates $\mM_1,\cdots,\mM_K$ given $\valpha_{b,n}$ from Eq. \eqref{eq:lcsa}:
\begin{align}
&\nonumber\\[-25pt]
& \text{\texttt{for i=1,...,dict\_iter:}}\nonumber\\[-2pt]
& \qquad\forall_k,\;\mM_k:=\mM_k\!-\!\lambda_{\text{DL}}\!\!\sum_{n=1}^{N\!\cdot\! Z+1}\!\!\nabla_{\mM_k}d^2_{\text{uDTW}^\bullet}\Big(\mPsi^\ddag_{b,n}, \sum_{l=1}^K\alpha_{l,b,n}\mM_{l}\Big),
\end{align}
where \texttt{dict\_iter} is set to 10 and $\lambda_{\text{DL}}\!=\!0.001$; 
\item The main loss for the Feature Encoder update step is given as ($\lambda_{\text{EN}}\!=\!0.001$):
\begin{align}
& \mathcal{P}:=\mathcal{P}\!-\!\lambda_{\text{EN}}\!\!\sum_{n=1}^{N\!\cdot\! Z+1}\!\!\nabla_{\mathcal{P}}d^2_{\text{uDTW}^\bullet}\Big(\mPsi^\ddag_{b,n}, \mM'\Big) \!+\!\beta\Omega_\bullet\Big(\mPsi^\ddag_{b,n}, \mM'\Big), \\[-6pt]
&\qquad\;\text{where}\;\; \mM'\!=\!\sum_{l=1}^K\alpha_{l,b,n}\mM_{l}.\nonumber
\end{align}
\end{enumerate}

During  testing, we  use the learnt dictionary, pass new support and query sequences via Eq. \eqref{eq:lcsa} and obtain $\valpha$ codes. Subsequently, we compare the LCSA code of the query sequence with LCSA codes of support sequences via the histogram intersection kernel. The closest match in the support set determines the test label of the query sequence.

\subsection{Time Series Forecasting and Classification}
One of key applications of DTW and sDTW is learning with time series, including forecasting the evolution of time series as in Figure~\ref{fig:forecast} and time series classification.

\paragraph{Forecasting the Evolution of Time Series.} Let $\vx\!\in\!\mbr{t}$ and $\vx'\!\in\!\mbr{\tau-t}$ be the training and testing parts of one time series corresponding to timesteps $1,\cdots,t$ and $t\!+\!1,\cdots,\tau$, respectively. The goal is to learn encoder $f(\vx; \mathcal{P})\!\in\!\mbr{\tau-t}$ which will be able to take $\vx$ as input, learn to translate it to $\vx'$. Figure \ref{fig:forecast} show the full pipeline. We took the Encoding Network from the original soft-DTW pipeline \cite{marco2017icml}. Our training objective is:
\begin{align}
&\nonumber\\[-17pt]
    &\argmin_{\mathcal{P}}\sum_{n\in\idx{N}}d^2_{\text{uDTW}^\bullet}(\vpsi_n,\vx'_{n})\!+\beta\Omega_\bullet(\vpsi_n,\vx'_{n}),\\[-21pt]
        &\nonumber
\end{align}
where $\vpsi\!=\!f(\vx;\mathcal{P})$ and $N$ is the number of training time series,
$\mathcal{P}\!\equiv\![\mathcal{P}_{MLP},\mathcal{P}_{SN}]$ is the  set of  parameters of EN and SigmaNet. 
In order to obtain $\cov$, vectors $\vpsi$ and $\vx'$ are passed via SigmaNet. After training, at the test time, for a previously unseen testing sample $\vx$, $f(\cdot)$ has to predict 
the remaining part of the time series given by $\vx'$.

%

\paragraph{Time Series Classification.} Below we follow the setting for this classical task according to the original soft-DTW paper \cite{marco2017icml}, and define the \textbf{nearest centroid} classifier.
We estimate the Fr\'echet mean of training time series of each class separately. We do not use any Encoding Network but the raw  features. Let $\vx\!\in\!\mbr{\tau}$ be training samples and $\vmu\!\in\!\mbr{\tau'}$ be class prototypes ($\tau'$ is set to average of $\tau$ across all classes). We have:
\begin{align}
&\nonumber\\[-17pt]
    &\forall_{c},\;\argmin_{\mathcal{P}}\sum_{n\in\idx{N_c}}d^2_{\text{uDTW}^\bullet}(\vx_n,\vmu_c)\!+\beta\Omega_\bullet(\vx_n,\vmu_c),\\[-21pt]
    &\nonumber
\end{align}
%
%
\noindent
where $N_c$ is the number of samples for class $c\!\in\!\idx{C}$ and $\mathcal{P}\!\equiv\![\mathcal{P}_{SN},\boldsymbol{\mu}_c]$. During testing,  we apply $\argmin_{c\in\idx{C}}d^2_{\text{uDTW}^\bullet}(\vx,\vmu_c)\!+\beta\Omega_\bullet(\vx,\vmu_c)$ for $\vx$ to find its nearest neighbor and label it. The variances of $\vx$ are recovered through SigmaNet while variances of $\vmu_c$ were obtained during training (adding both yields $\cov$ of testing sample). %
As in soft-DTW paper \cite{marco2017icml}, we  use uDTW to directly find the \textbf{nearest neighbor} of $\vx$ across  training samples to label $\vx$ (for uncertainty, we use SigmaNet from the nearest centroid task).

\section{Experiments}
Below we apply uDTW in several scenarios such as (i) forecasting the evolution of time series, (ii) clustering/classifying time series, (iii) supervised few-shot action recognition, and (iv) unsupervised  few-shot action recognition.
%
%
%

\vspace{0.05cm}
\noindent\textbf{Datasets.} The following datasets are used in our experiments: 
\renewcommand{\labelenumi}{\roman{enumi}.}
\begin{enumerate}[leftmargin=0.6cm]
\item{{\em UCR}} archive~\cite{UCRArchive} is a dataset for time series classification archive. This dataset contains a wide variety of fields (astronomy, geology, medical imaging) and lengths, and can be used for time series classification/clustering and forecasting tasks.

\item{{\em NTU RGB+D (NTU-60)}}~\cite{Shahroudy_2016_NTURGBD} contains 56,880 video sequences and over 4 million frames. 
NTU-60 has variable sequence lengths  and  high intra-class variations. 

\item{{\em NTU RGB+D 120 (NTU-120)}}~\cite{Liu_2019_NTURGBD120}, an extension of NTU-60, contains 120 action classes (daily/health-related), and 114,480 RGB+D video samples  captured with 106 distinct human subjects from 155 different camera viewpoints.

\item{{\em Kinetics}}~\cite{kay2017kinetics} is a large-scale collection of 650,000 video clips that cover 400/600/700 human action classes.
It includes human-object interactions such as {\it playing instruments}, as well as human-human interactions such as {\it shaking hands} and {\it hugging}. 
We follow approach~\cite{stgcn2018aaai} and use the estimated joint locations in the pixel coordinate system as the input to our pipeline.
As OpenPose  produces the 2D body joint coordinates and Kinetics-400 does not offer multiview or depth data, we use a network of Martinez et al.   \cite{martinez_2d23d} pre-trained on Human3.6M~\cite{Catalin2014Human3}, combined with the 2D OpenPose output to  estimate 3D coordinates from 2D coordinates. The 2D OpenPose and the latter network give us $(x,y)$ and $z$ coordinates, respectively.
\end{enumerate}

\subsection{Fr\'echet Mean of Time Series}

Below, we visually inspect the Fr\'echet mean for the Euclidean, sDTW and our uDTW distance, respectively. 

\begin{figure}[t]
\centering
\begin{subfigure}[b]{0.45\linewidth}
\includegraphics[trim=1cm 1cm 1cm 2.5cm, clip=true,width=0.99\linewidth]{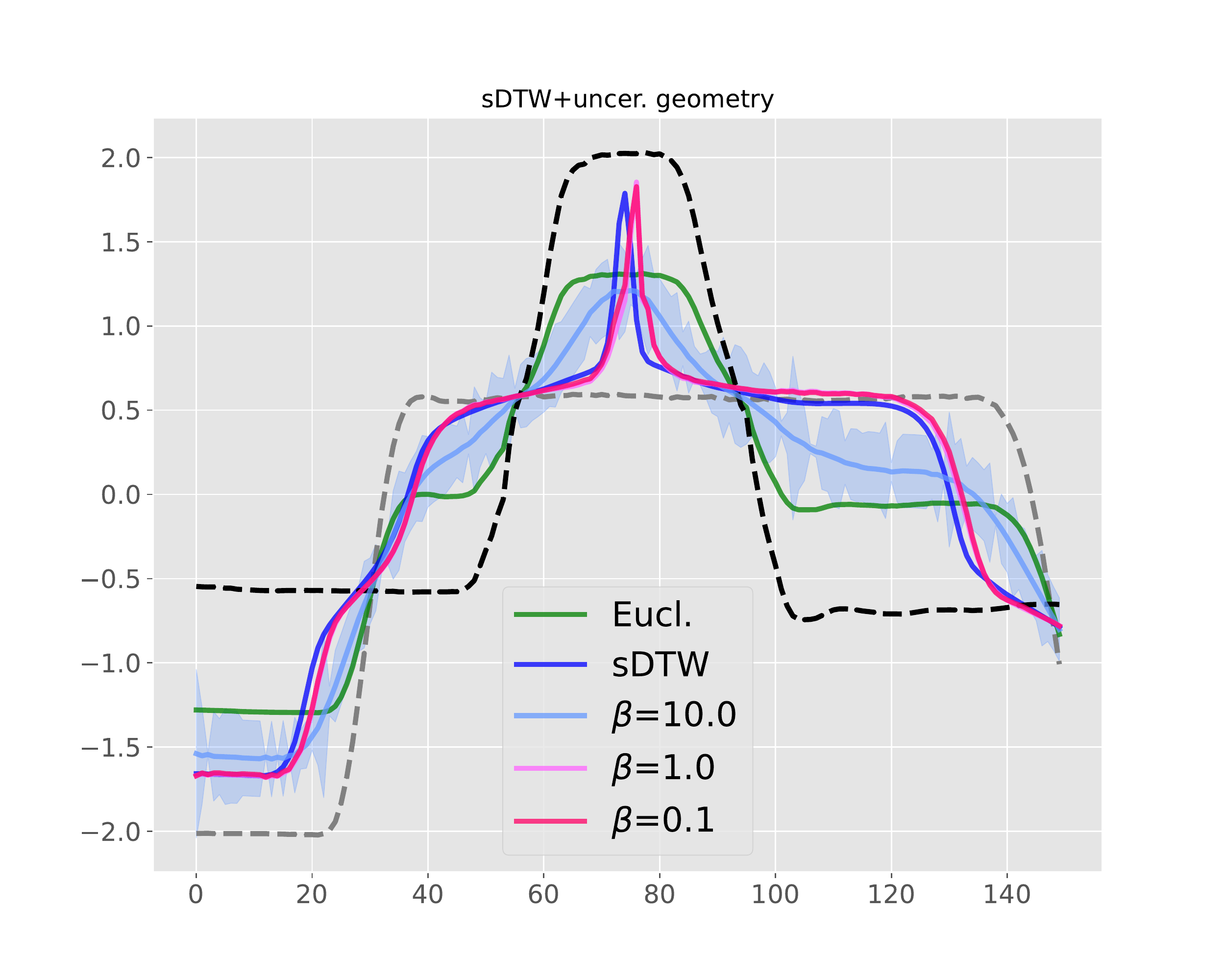}
\caption{$\beta$ (where $\lambda\!=\!1$)}\label{fig:beta}
\end{subfigure}
\begin{subfigure}[b]{0.45\linewidth}
\includegraphics[trim=1cm 1cm 1cm 2.5cm, clip=true,width=0.99\linewidth]{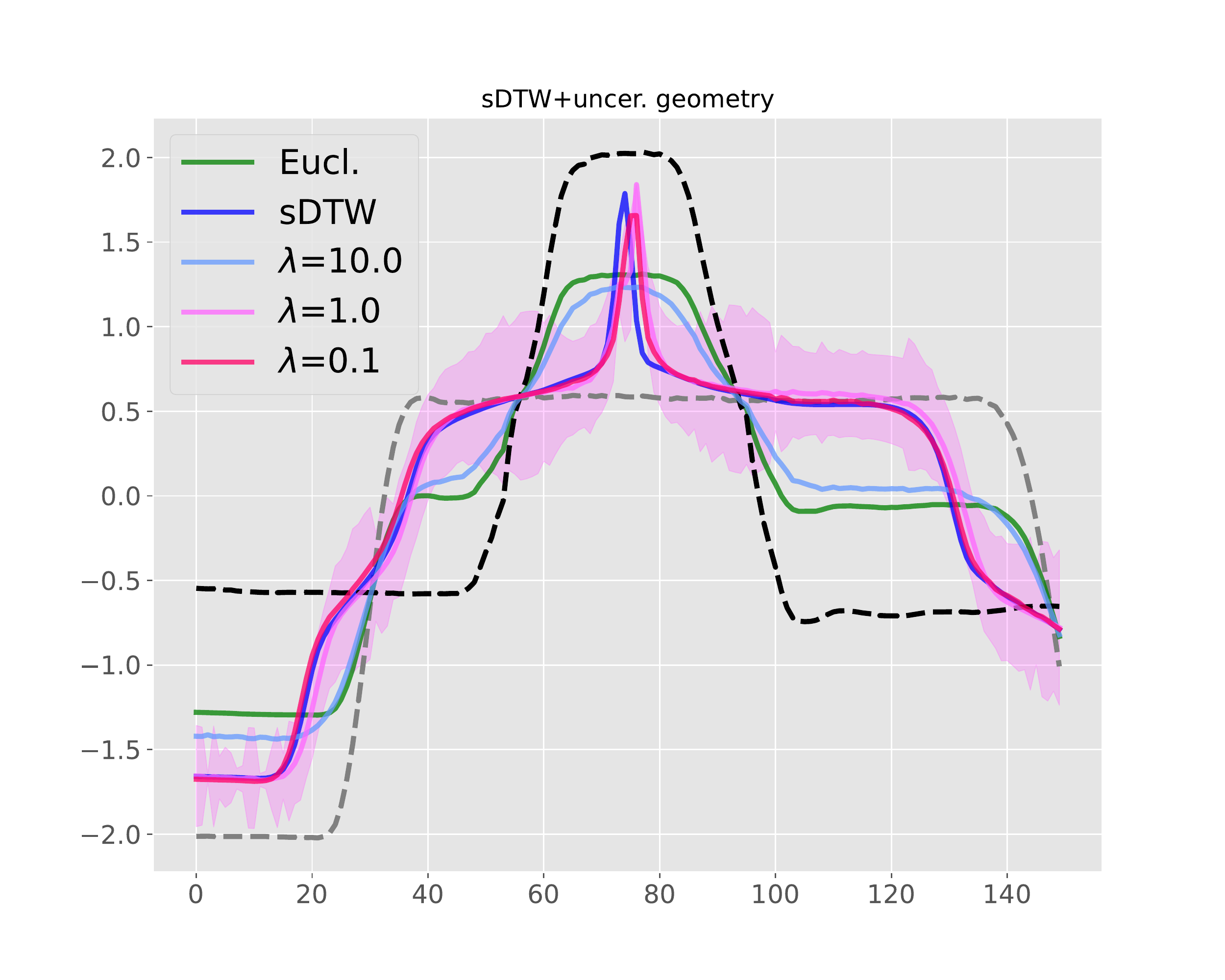}
\caption{$\lambda$ (where $\beta\!=\!10$)}\label{fig:lambda}
\end{subfigure}
\caption{Interpolation between two time series (grey and black dashed lines) 
on the Gun Point dataset. 
We compute the barycenter by solving 
%
$\argmin\limits_{\vmu, \vsigma_\mu}\sum_{n=1}^{2}
d^2_{\text{uDTW}}\big( \mD, \cov^\dag\big)+\beta\Omega\big(\cov\big)+\lambda\Omega'\big(\cov\big)$ where  $\mD\!=\!(\vx_n\mathbf{1}^\top\!-\!\mathbf{1}\vmu^\top)^2$ and  $\cov\!=\!\mathbf{1}\mathbf{1}^\top\!\!+\!\mathbf{1}\vsigma_\mu^\top$
%
%
%
where $\vx_n$ is the given $n$-th time series. 
$\beta\!\geq\!0$ controls the penalty for high matching uncertainty,
$\Omega'$ is defined as in Eq. \eqref{eq:unc_pen} but element-wise $\log\!\cov$ is replaced by element-wise $(\cov\!-\!1)^2$ so that
 $\lambda\!\geq\!0$ favours uncertainty to remain close to one. 
 $\beta$ and $\lambda$ control the uncertainty estimation and yield  different barycenters than  the Euclidean (green color) and sDTW (blue color) distances. As $\Omega$ and $\Omega'$ act similar, we only use $\Omega$ in our experiments.
%
}
\label{fig:interp}
\end{figure}

\begin{figure*}[t]
\centering
\begin{subfigure}[t]{0.9\linewidth}
\includegraphics[trim=4cm 1.2cm 3.5cm 0.6cm, clip=true,width=\linewidth]{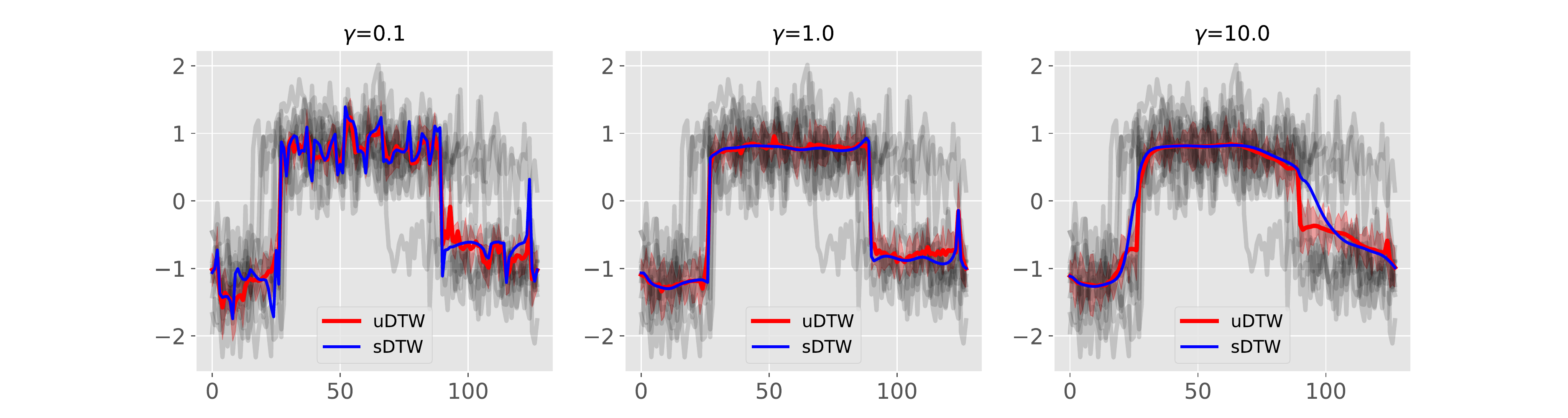}
\caption{CBF}\label{fig:cbf}
\end{subfigure}\\
%

\begin{subfigure}[t]{0.9\linewidth}
\includegraphics[trim=4cm 1.2cm 3.5cm 0.6cm, clip=true,width=\linewidth]{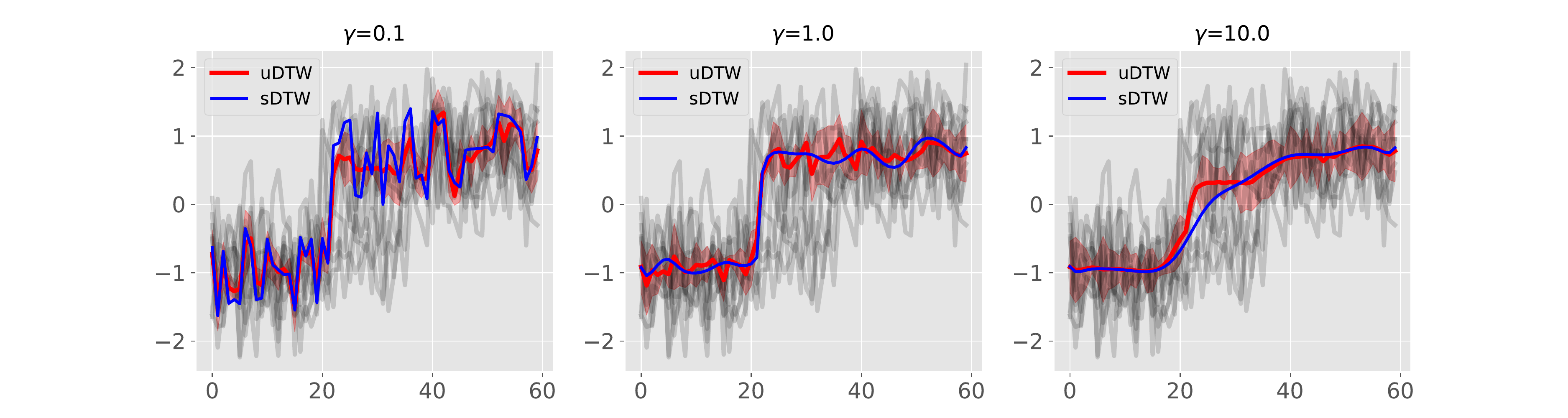}
\caption{Synthetic Control}\label{fig:syn}
\end{subfigure}

\caption{Comparison of barycenter based on sDTW or uDTW on CBF 
and Synthetic Control. 
We visualize uncertainty around the barycenters in red color for uDTW. Our uDTW generates reasonable barycenters even when higher $\gamma$ values are used, \eg, $\gamma\!=\!10.0$. Higher $\gamma$ value leads to smooth barycenter but introducing higher uncertainty. 
}
\label{fig:barycenter}
\end{figure*}

\paragraph{Experimental setup.} We follow  the protocol of soft-DTW paper \cite{marco2017icml}. For each dataset in UCR, we choose a class at random, pick 10 time series from the selected class to compute its barycenter. We use L-BFGS~\cite{lbfgs} to minimise the proposed uDTW barycenter objective. We set the maximum number of iterations to 100.

\paragraph{Qualitative results.} We first perform averaging between two time series (Figure~\ref{fig:interp}). We notice that averaging under the uDTW yields substantially different results than those obtained with the Euclidean and sDTW geometry. 

Figure~\ref{fig:barycenter} shows the barycenters obtained using sDTW and our uDTW. We observe that our uDTW yields more reasonable barycenters than sDTW even when large $\gamma$ are used, \eg, for $\gamma\!=\!10$ (right column of plots in Figure~\ref{fig:barycenter}), the change points of red curve look sharper. We also notice that both uDTW and sDTW with low smoothing parameter $\gamma\!=\!0.1$ can get stuck in some bad local minima, but our uDTW has fewer sharp peaks compared with sDTW (barycenters of uDTW are improved by the uncertainty measure). Moreover, higher $\gamma$ values smooth the barycenter but introducing higher uncertainty (see uncertainty visualization around the barycenters by comparing, \eg, $\gamma\!=\!0.1$ \vs $\gamma\!=\!10.0$). With $\gamma\!=\!1$, the barycenters of sDTW and uDTW match well with the time series. More visualizations can be found in Appendix Sec.~\ref{supp:barycenter}.

\subsection{Classification of Time Series}

In this section, we devise the nearest neighbor and nearest centroid classifiers~\cite{hastie01statisticallearning} with uDTW, as detailed in Section \ref{sec:approach}.  
%
%
For the $K$-nearest neighbor classifier, we used softmax for the final decision.
~See Appendix Sec. \ref{sec:softmax} for details.


\paragraph{Experimental setup.} We use 50\% of the data for training, 25\% for validation and 25\% for testing. We report $K=$ 1, 2 and 3 for the nearest neighbor classifier.

\paragraph{Quantitative results.} Table~\ref{tab:ucrclassification} shows a comparison of our uDTW versus Euclidean, DTW, sDTW, and sDTW div. Unsurprisingly, the use of uDTW for barycenter computation improves the accuracy of the nearest centroid classifier, and it outperforms sDTW div. by $\sim$ 2\%. Moreover, uDTW  boosts results for the nearest neighbor classifier given $K$=1, 2 and 3 by 1.4\%, 1.7\% and 3.2\%, respectively, compared to sDTW div.

 
%

\begin{table}[tbp]
\caption{Classification accuracy (mean$\pm$std) on UCR archive by the nearest neighbor and the nearest centroid classifiers. In the column we indicate which distance was used for computing the class prototypes. $K$ is the number of nearest neighbors in this context.}
\label{tab:ucrclassification}
\begin{center}
\begin{tabular}{lcccc}
\toprule
 & \multicolumn{3}{c}{\bf Nearest neighbor} &{\bf Nearest centroid}\\
 \cline{2-4}
 
 & $K = 1$ & $K = 3$ & $K = 5$ & \\
\midrule
Euclidean & 71.2$\pm$17.5& 72.3$\pm$18.1& 73.0$\pm$16.7&61.3$\pm$20.1\\
DTW~\cite{marco2011icml} & 74.2$\pm$16.6&75.0$\pm$17.0&75.4$\pm$15.8&65.9$\pm$18.8 \\
sDTW~\cite{marco2017icml} & 76.2$\pm$16.6&77.2$\pm$15.9& 78.0$\pm$16.5&70.5$\pm$17.6\\ 
sDTW div.~\cite{pmlr-v130-blondel21a}& 78.6$\pm$16.2& 79.5$\pm$16.7&80.1$\pm$16.5\!\!& 70.9$\pm$17.8\\ 
\hdashline
\rowcolor{blue!10}
uDTW & 80.0$\pm$15.0& 81.2$\pm$17.8&83.3$\pm$16.2& 72.2$\pm$16.0\\
\bottomrule
\end{tabular}
\end{center}
\end{table}

\subsection{Forecasting the Evolution of Time Series}
\noindent{\bf Experimental setup.} We use the training and test sets pre-defined in the UCR archive. For both training and test, we use the first 60\% of timesteps of series as input and the remaining 40\% as output, ignoring the class information. 

\begin{figure}[t]
\centering
\begin{subfigure}[b]{0.45\linewidth}
\includegraphics[trim=0.6cm 9.2cm 12.6cm 0cm, clip=true,width=0.99\linewidth]{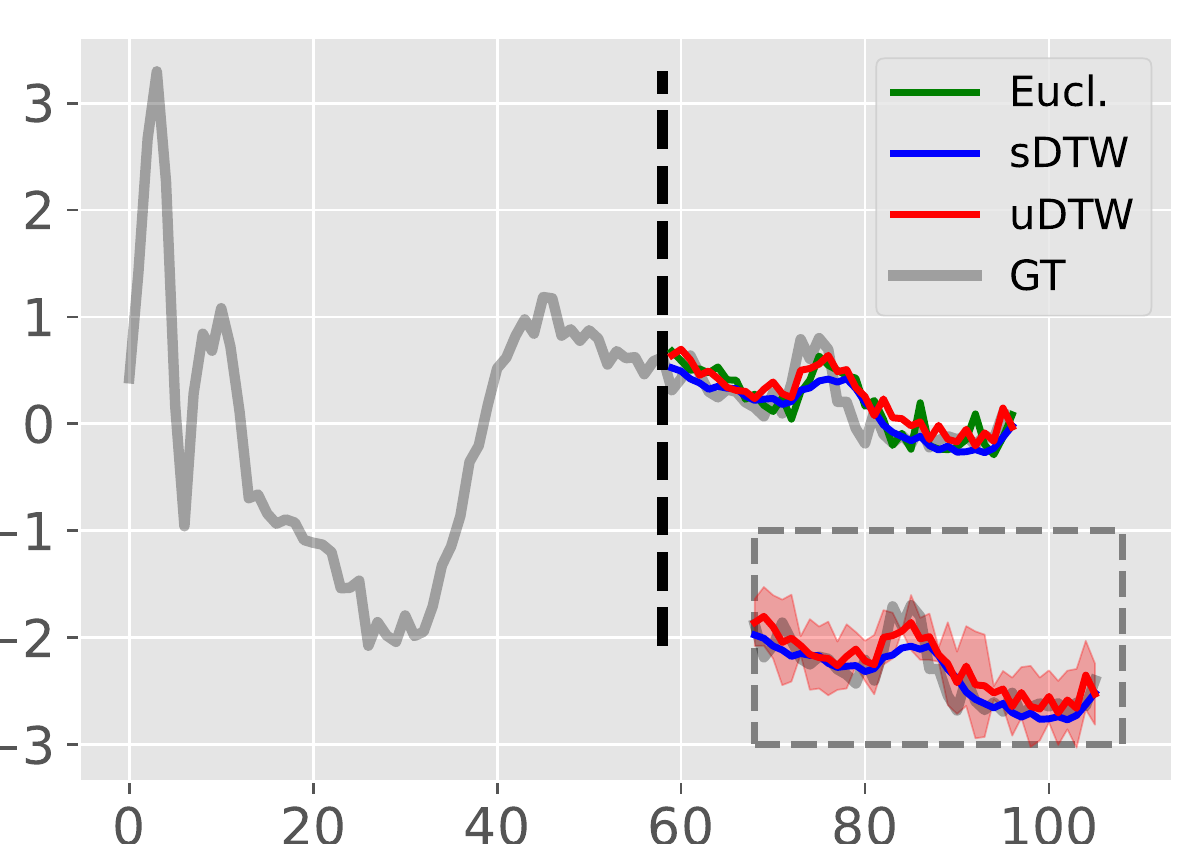}
\caption{ECG200}\label{fig:ecg200_pred}
\end{subfigure}
\begin{subfigure}[b]{0.45\linewidth}
\includegraphics[trim=12.6cm 0.5cm 0.6cm 9cm, clip=true,width=0.99\linewidth]{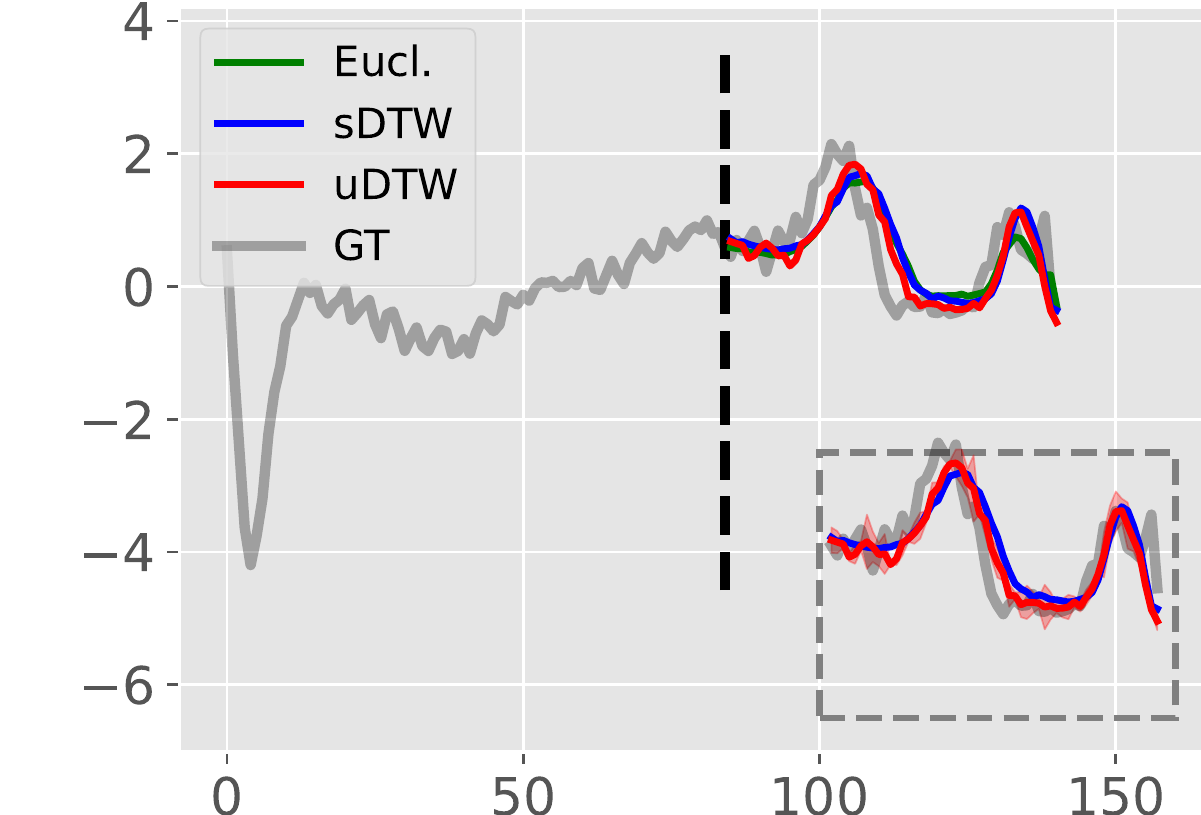}
\caption{ECG5000}\label{fig:ecg5000_pred}
\end{subfigure}
\caption{Given the first part of a time series, we train 3 multi-layer perception (MLP) to predict the remaining part, we use the Euclidean, sDTW or uDTW distance per MLP. We use ECG200 and ECG5000 in UCR archive,  and display  the prediction obtained for the given test sample with either of these 3 distances and the ground truth (GT). Oftentimes, we observe that uDTW  helps predict the sudden changes well.}
\label{fig:pred}
\end{figure}

\paragraph{Qualitative results.} The visualization of the predictions are given in Figure~\ref{fig:pred}. Although the predictions under the sDTW and uDTW losses sometimes agree with each other, they can be visibly different. Predictions under uDTW can confidently predict the abrupt and sharp changes. More visualizations can be found in Appendix Sec.~\ref{supp:pred}.

\paragraph{Quantitative results.} We also provide quantitive results to validate the effectiveness of uDTW. We use ECG5000 dataset from the UCR archive which is composed of 5000 electrocardiograms (ECG) (500 for training and 4500 for testing) of length 140. To better evaluate the predictions, we use 2 different metrics (i) MSE for the predicted errors of each time step (ii) DTW, sDTW div. and uDTW for comparing the `shape' of time series. We use such  shape metrics for evaluation as the length of time series generally varies, and the MSE metric may lead to  biased results which ignore the shape trend of time series. We then use the Student's $t$-test (with significance level 0.05) to highlight the best performance in each experiment (averaged over 100 runs). Table~\ref{tab:pred} shows that our uDTW achieves almost the best performance on both MSE and shape evaluation metrics (lower score is better).

\begin{table}[t]
\caption{Time series forecasting results evaluated with MSE, DTW, sDTW div. and uDTW metrics on ECG5000, averaged over 100 runs (mean$\pm$std). 
Best method(s) are highlighted in bold using Student's $t$-test.
Column-wise distances indicate the distance used during training. Row-wise distances indicate the distance used to compare prediction with the groundtruth at the test time (lower values are better). 
}
\label{tab:pred}
\begin{center}
\begin{tabular}{lcccc}
\toprule
&  MSE & DTW & sDTW div. & uDTW\\
\hline
Euclidean & {\bf 32.1$\pm$1.62}&20.0$\pm$0.18& 15.3$\pm$0.16 & 14.4$\pm$0.18\\
sDTW~\cite{marco2017icml} &38.6$\pm$6.30&{\bf 17.2$\pm$0.80}&22.6$\pm$3.59& 32.1$\pm$2.25\\
sDTW div.~\cite{pmlr-v130-blondel21a} &24.6$\pm$1.37&38.9$\pm$5.33&{\bf 20.0$\pm$2.44}& 15.4$\pm$1.62\\
\rowcolor{blue!10}
uDTW &23.0$\pm$1.22&{\bf 16.7$\pm$0.08}& {\bf 16.8$\pm$1.62}&{\bf 8.27$\pm$0.79} \\
\bottomrule
\end{tabular}
\end{center}
\end{table}
\subsection{Few-shot Action Recognition}

Below, we use uDTW as a distance in our objectives for few-shot action recognition (AR) tasks. We implement  supervised and unsupervised pipelines (which is also novel). 

\paragraph{Experimental setup.} For NTU-120, we follow the standard one-shot protocols~\cite{Liu_2019_NTURGBD120}. Base on this protocol, we create a similar one-shot protocol for NTU-60, with 50/10 action classes used for training/testing respectively (see Appendix Sec.~\ref{app:epr} for details). We also evaluate the model on both 2D and 3D Kinetics-skeleton. We split the whole Kinetics-skeleton into 200 actions for training (the rest is used for testing).  
We choose Matching Nets (MatchNets) and Prototypical Net (ProtoNet) as baselines as these two models are very popular baselines, 
and we adapt these methods to skeleton-based action recognition. We reshape and resize each video block into 224$\times$224 color image, and pass this image into MatchNets and ProtoNet to learn the feature representation per video block. We compare uDTW \vs Euclidean, sDTW, sDTW div. and recent TAP.

\paragraph{Quantitative results.} Table~\ref{tab:ntu60},~\ref{tab:ntu120} and~\ref{tab:kinetics} show that our uDTW performs better than sDTW and sDTW div. on both supervised and unsupervised few-shot action recognition. On Kinetics-skeleton dataset, we gain 2.4\% and 4.4\% improvements on 3D skeletons for supervised and unsupervised settings. On supervised setting, we outperform TAP by $\sim$ 4\% and 2\% on NTU-60 and NTU-120 respectively. Moreover, we outperform sDTW by $\sim$ 2\% and 3\% on NTU-60 and NTU-120 for the unsupervised setting. 
More evaluations on few-shot action recognition are in Appendix Sec.~\ref{supp:fsl}.

\begin{table}[tbp]
\begin{minipage}{.5\linewidth}
\caption{Evaluations on NTU-60. }
\label{tab:ntu60}
\centering
\begin{tabular}{lccccc}
\toprule
\#classes & 10 & 20 & 30 & 40 & 50\\
\midrule
& \multicolumn{5}{c}{\bf Supervised}\\
\hline
MatchNets~\cite{f4Matching} 
& 46.1 & 48.6 & 53.3 & 56.3 & 58.8 \\
ProtoNet~\cite{f1} 
& 47.2 & 51.1 & 54.3 & 58.9 & 63.0 \\
TAP~\cite{su2022temporal}
& 54.2 & 57.3 & 61.7 & 64.7 & 68.3 \\
\hdashline
Euclidean & 38.5 & 42.2 & 45.1 & 48.3 & 50.9\\
sDTW~\cite{marco2017icml} & 53.7 & 56.2 & 60.0 & 63.9 & 67.8\\
sDTW div.~\cite{pmlr-v130-blondel21a} & 54.0 & 57.3 & 62.1 & 65.7 & 69.0\\
\rowcolor{blue!10}
uDTW & 56.9 & 61.2 & 64.8 & 68.3 & 72.4\\
\hline
& \multicolumn{5}{c}{\bf Unsupervised}\\
\hline
Euclidean & 20.9 & 23.7 & 26.3 & 30.0 & 33.1\\
sDTW~\cite{marco2017icml} & 35.6 & 45.2 & 53.3 & 56.7 & 61.7\\
sDTW div.~\cite{pmlr-v130-blondel21a} & 36.0 & 46.1 & 54.0 & 57.2 & 62.0\\
\rowcolor{blue!10}
uDTW & 37.0 & 48.3 & 55.3 & 58.0 & 63.3\\
\bottomrule
\end{tabular}
\end{minipage}\hfill
\begin{minipage}{.5\linewidth}
\caption{Evaluations on NTU-120.}
\label{tab:ntu120}
\centering
\begin{tabular}{lccccc}
\toprule
\#classes & 20 & 40 & 60 & 80 & 100\\
\midrule
& \multicolumn{5}{c}{\bf Supervised}\\
\hline
MatchNets~\cite{f4Matching} 
& 20.5 & 23.4 & 25.1 & 28.7 & 30.0 \\
ProtoNet~\cite{f1} 
& 21.7 & 24.0 & 25.9 & 29.2 & 32.1 \\
TAP~\cite{su2022temporal}
& 31.2 & 37.7 & 40.9 & 44.5 &  47.3\\
\hdashline
Euclidean & 18.7 & 21.3 & 24.9 & 27.5 & 30.0\\
sDTW~\cite{marco2017icml} & 30.3 & 37.2 & 39.7 & 44.0 & 46.8\\
sDTW div.~\cite{pmlr-v130-blondel21a} & 30.8 & 38.1 & 40.0 & 44.7 & 47.3\\
\rowcolor{blue!10}
uDTW & 32.2 & 39.0 & 41.2 & 45.3 & 49.0\\
\hline
& \multicolumn{5}{c}{\bf Unsupervised}\\
\hline
Euclidean & 13.5 & 16.3 & 20.0 & 24.9 & 26.2\\
sDTW~\cite{marco2017icml} & 20.1 & 25.3 & 32.0 & 36.9 & 40.9\\
sDTW div.~\cite{pmlr-v130-blondel21a} & 20.8 & 26.0 & 33.2 & 37.5 & 42.3\\
\rowcolor{blue!10}
uDTW & 22.7 & 28.3 & 35.9 & 39.4 & 44.0\\
\bottomrule
\end{tabular}
\end{minipage}
\end{table}


\begin{table}[tbp]
\caption{Evaluations on 2D and 3D Kinetics-skeleton.}
\label{tab:kinetics}
\begin{center}
\begin{tabular}{lcccc}
\toprule
& \multicolumn{2}{c}{\bf Supervised} & \multicolumn{2}{c}{\bf Unsupervised}\\
\cline{2-5}
& 2D & 3D & 2D & 3D\\
\midrule
Euclidean & 21.2 & 23.1 & 12.7 & 13.3\\
TAP~\cite{su2022temporal}
& 32.9 & 36.0 & - & -\\
sDTW~\cite{marco2017icml}  & 34.7 & 39.6 & 23.3& 28.3\\
sDTW div.~\cite{pmlr-v130-blondel21a}  & 35.0 & 40.1 & 24.0& 28.9\\
\rowcolor{blue!10}
uDTW & 35.5 & 42.0 & 25.9 & 32.7\\
\bottomrule
\end{tabular}
\end{center}
\end{table}

\section{Conclusions}

We have introduced the uncertainty-DTW which handles the uncertainty estimation of frame- and/or block-wise features to improve the path warping of the celebrated soft-DTW. Our uDTW  produces the uncertainty-weighted distance along the path and returns the regularization penalty aggregated along the path, which follows sound principles of classifier regularization. We have provided several pipelines for time series forecasting, and supervised and unsupervised action recognition, which use uDTW as a distance. Our simple  uDTW achieves better sequence alignment in several benchmarks. 

%
%
\bibliographystyle{splncs04}
\bibliography{egbib}

\begin{thebibliography}{10}
\providecommand{\url}[1]{\texttt{#1}}
\providecommand{\urlprefix}{URL }
\providecommand{\doi}[1]{https://doi.org/#1}

\bibitem{10.5555/3327546.3327715}
Abid, A., Zou, J.: Autowarp: Learning a warping distance from unlabeled time
  series using sequence autoencoders. NIPS'18, Curran Associates Inc., Red
  Hook, NY, USA (2018)

\bibitem{9523103}
Ben-Ari, R., Shpigel~Nacson, M., Azulai, O., Barzelay, U., Rotman, D.: Taen:
  Temporal aware embedding network for few-shot action recognition. In: 2021
  IEEE/CVF Conference on Computer Vision and Pattern Recognition Workshops
  (CVPRW). pp. 2780--2788 (2021)

\bibitem{pmlr-v130-blondel21a}
Blondel, M., Mensch, A., Vert, J.P.: Differentiable divergences between time
  series. In: Banerjee, A., Fukumizu, K. (eds.) Proceedings of The 24th
  International Conference on Artificial Intelligence and Statistics.
  Proceedings of Machine Learning Research, vol.~130, pp. 3853--3861. PMLR
  (13--15 Apr 2021)

\bibitem{kaidi2020cvpr}
Cao, K., Ji, J., Cao, Z., Chang, C.Y., Niebles, J.C.: Few-shot video
  classification via temporal alignment. In: CVPR (2020)

\bibitem{Catalin2014Human3}
Catalin, Ionescu, Dragos, Papava, Vlad, Olaru, Cristian, Sminchisescu:
  Human3.6m: Large scale datasets and predictive methods for 3d human sensing
  in natural environments. IEEE Transactions on Pattern Analysis \& Machine
  Intelligence  (2014)

\bibitem{marco2011icml}
Cuturi, M.: Fast global alignment kernels. In: International Conference on
  Machine Learning (ICML) (2011)

\bibitem{marco2017icml}
Cuturi, M., Blondel, M.: Soft-dtw: a differentiable loss function for
  time-series. In: International Con- ference on Machine Learning (ICML) (2017)

\bibitem{UCRArchive}
Dau, H.A., Keogh, E., Kamgar, K., Yeh, C.C.M., Zhu, Y., Gharghabi, S.,
  Ratanamahatana, C.A., Yanping, Hu, B., Begum, N., Bagnall, A., Mueen, A.,
  Batista, G.: {The UCR Time Series Classification Archive} (October 2018),
  \url{https://www.cs.ucr.edu/~eamonn/time\_series\_data_2018/}

\bibitem{10.1145/3447548.3467231}
Dempster, A., Schmidt, D.F., Webb, G.I.: Minirocket: A very fast (almost)
  deterministic transform for time series classification. In: Proceedings of
  the 27th ACM SIGKDD Conference on Knowledge Discovery \& Data Mining. p.
  248–257. KDD '21, Association for Computing Machinery, New York, NY, USA
  (2021). \doi{10.1145/3447548.3467231}

\bibitem{donahue2021endtoend}
Donahue, J., Dieleman, S., Binkowski, M., Elsen, E., Simonyan, K.: End-to-end
  adversarial text-to-speech. In: International Conference on Learning
  Representations (2021)

\bibitem{dosovitskiy2020image}
Dosovitskiy, A., Beyer, L., Kolesnikov, A., Weissenborn, D., Zhai, X.,
  Unterthiner, T., Dehghani, M., Minderer, M., Heigold, G., Gelly, S., et~al.:
  An image is worth 16x16 words: Transformers for image recognition at scale.
  In: International Conference on Learning Representations (2020)

\bibitem{4668349}
García-García, D., Parrado~Hernández, E., Díaz-de María, F.: A new
  distance measure for model-based sequence clustering. IEEE Transactions on
  Pattern Analysis and Machine Intelligence  \textbf{31}(7),  1325--1331
  (2009). \doi{10.1109/TPAMI.2008.268}

\bibitem{hastie01statisticallearning}
Hastie, T., Tibshirani, R., Friedman, J.: The Elements of Statistical Learning.
  Springer Series in Statistics, Springer New York Inc., New York, NY, USA
  (2001)

\bibitem{uncertainty4}
H{\"{u}}llermeier, E., Waegeman, W.: Aleatoric and epistemic uncertainty in
  machine learning: an introduction to concepts and methods. Mach. Learn.
  \textbf{110}(3),  457--506 (2021). \doi{10.1007/s10994-021-05946-3}

\bibitem{uncertainty3}
Indrayan, A.: Medical biostatistics. Chapman \& Hall/CRC,, Boca Raton :, 2nd
  ed. edn. (c2008),
  \url{http://www.loc.gov/catdir/toc/ecip0723/2007030353.html}

\bibitem{kay2017kinetics}
Kay, W., Carreira, J., Simonyan, K., Zhang, B., Hillier, C., Vijayanarasimhan,
  S., Viola, F., Green, T., Back, T., Natsev, P., Suleyman, M., Zisserman, A.:
  The kinetics human action video dataset (2017)

\bibitem{uncertainty5}
Kendall, A., Gal, Y.: What uncertainties do we need in bayesian deep learning
  for computer vision? In: Guyon, I., Luxburg, U.V., Bengio, S., Wallach, H.,
  Fergus, R., Vishwanathan, S., Garnett, R. (eds.) Advances in Neural
  Information Processing Systems. vol.~30. Curran Associates, Inc. (2017)

\bibitem{uncertainty2}
Kiureghian, A.D., Ditlevsen, O.: Aleatory or epistemic? does it matter?
  Structural Safety  \textbf{31}(2),  105--112 (2009).
  \doi{https://doi.org/10.1016/j.strusafe.2008.06.020}, risk Acceptance and
  Risk Communication

\bibitem{6116129}
Koniusz, P., Mikolajczyk, K.: Soft assignment of visual words as linear
  coordinate coding and optimisation of its reconstruction error. In: 2011 18th
  IEEE International Conference on Image Processing. pp. 2413--2416 (2011).
  \doi{10.1109/ICIP.2011.6116129}

\bibitem{hosvd}
Koniusz, P., Wang, L., Cherian, A.: Tensor representations for action
  recognition. TPAMI  (2020)

\bibitem{KONIUSZ2013479}
Koniusz, P., Yan, F., Mikolajczyk, K.: Comparison of mid-level feature coding
  approaches and pooling strategies in visual concept detection. Computer
  Vision and Image Understanding  \textbf{117}(5),  479--492 (2013).
  \doi{https://doi.org/10.1016/j.cviu.2012.10.010}

\bibitem{ta2n2021}
Li, S., Liu, H., Qian, R., Li, Y., See, J., Fei, M., Yu, X., Lin, W.: {TTAN:}
  two-stage temporal alignment network for few-shot action recognition. CoRR
  (2021)

\bibitem{lbfgs}
Liu, D.C., Nocedal, J.: On the limited memory bfgs method for large scale
  optimization. Mathematical Programming  \textbf{45},  503–528 (1989)

\bibitem{Liu_2019_NTURGBD120}
Liu, J., Shahroudy, A., Perez, M., Wang, G., Duan, L.Y., Kot, A.C.: Ntu rgb+d
  120: A large-scale benchmark for 3d human activity understanding. IEEE
  Transactions on Pattern Analysis and Machine Intelligence  (2019).
  \doi{10.1109/TPAMI.2019.2916873}

\bibitem{6126534}
Liu, L., Wang, L., Liu, X.: In defense of soft-assignment coding. In: 2011
  International Conference on Computer Vision. pp. 2486--2493 (2011).
  \doi{10.1109/ICCV.2011.6126534}

\bibitem{Lohit_2019_CVPR}
Lohit, S., Wang, Q., Turaga, P.: Temporal transformer networks: Joint learning
  of invariant and discriminative time warping. In: Proceedings of the IEEE/CVF
  Conference on Computer Vision and Pattern Recognition (CVPR) (June 2019)

\bibitem{martinez_2d23d}
{Martinez}, J., {Hossain}, R., {Romero}, J., {Little}, J.J.: A simple yet
  effective baseline for 3d human pose estimation. In: 2017 IEEE International
  Conference on Computer Vision (ICCV). pp. 2659--2668 (2017).
  \doi{10.1109/ICCV.2017.288}

\bibitem{uncertainty1}
Matthies, H.G.: Quantifying uncertainty: Modern computational representation of
  probability and applications. In: Ibrahimbegovic, A., Kozar, I. (eds.)
  Extreme Man-Made and Natural Hazards in Dynamics of Structures. pp. 105--135.
  Springer Netherlands, Dordrecht (2007)

\bibitem{memmesheimer2021skeletondml}
Memmesheimer, R., Häring, S., Theisen, N., Paulus, D.: Skeleton-dml: Deep
  metric learning for skeleton-based one-shot action recognition (2021)

\bibitem{2021dml}
Memmesheimer, R., Theisen, N., Paulus, D.: Signal level deep metric learning
  for multimodal one-shot action recognition (2020)

\bibitem{pmlr-v80-mensch18a}
Mensch, A., Blondel, M.: Differentiable dynamic programming for structured
  prediction and attention. In: Dy, J., Krause, A. (eds.) Proceedings of the
  35th International Conference on Machine Learning. Proceedings of Machine
  Learning Research, vol.~80, pp. 3462--3471. PMLR (10--15 Jul 2018)

\bibitem{tarn_BMVC2019}
Mina, B., Zoumpourlis, G., Patras, I.: Tarn: Temporal attentive relation
  network for few-shot and zero-shot action recognition. In: Sidorov, K.,
  Hicks, Y. (eds.) Proceedings of the British Machine Vision Conference (BMVC).
  pp. 130.1--130.14. BMVA Press (September 2019). \doi{10.5244/C.33.130}

\bibitem{Perrett_2021_CVPR}
Perrett, T., Masullo, A., Burghardt, T., Mirmehdi, M., Damen, D.:
  Temporal-relational crosstransformers for few-shot action recognition. In:
  Proceedings of the IEEE/CVF Conference on Computer Vision and Pattern
  Recognition (CVPR). pp. 475--484 (June 2021)

\bibitem{ramachandran2018unsupervised}
Ramachandran, P., Liu, P.J., Le, Q.V.: Unsupervised pretraining for sequence to
  sequence learning (2018)

\bibitem{1163055}
Sakoe, H., Chiba, S.: Dynamic programming algorithm optimization for spoken
  word recognition. IEEE Transactions on Acoustics, Speech, and Signal
  Processing  \textbf{26}(1),  43--49 (1978). \doi{10.1109/TASSP.1978.1163055}

\bibitem{Shahroudy_2016_NTURGBD}
Shahroudy, A., Liu, J., Ng, T.T., Wang, G.: Ntu rgb+d: A large scale dataset
  for 3d human activity analysis. In: IEEE Conference on Computer Vision and
  Pattern Recognition (June 2016)

\bibitem{f1}
Snell, J., Swersky, K., Zemel, R.S.: Prototypical networks for few-shot
  learning. In: Guyon, I., von Luxburg, U., Bengio, S., Wallach, H.M., Fergus,
  R., Vishwanathan, S.V.N., Garnett, R. (eds.) Advances in Neural Information
  Processing Systems 30: Annual Conference on Neural Information Processing
  Systems 2017, 4-9 December 2017, Long Beach, CA, {USA}. pp. 4077--4087 (2017)

\bibitem{8466003}
Su, B., Hua, G.: Order-preserving optimal transport for distances between
  sequences. IEEE Transactions on Pattern Analysis and Machine Intelligence
  \textbf{41}(12),  2961--2974 (2019). \doi{10.1109/TPAMI.2018.2870154}

\bibitem{su2022temporal}
Su, B., Wen, J.R.: Temporal alignment prediction for supervised representation
  learning and few-shot sequence classification. In: International Conference
  on Learning Representations (2022)

\bibitem{9008561}
Su, B., Zhou, J., Wu, Y.: Order-preserving wasserstein discriminant analysis.
  In: 2019 IEEE/CVF International Conference on Computer Vision (ICCV). pp.
  9884--9893 (2019). \doi{10.1109/ICCV.2019.00998}

\bibitem{8851694}
Tan, S., Yang, R.: Learning similarity: Feature-aligning network for few-shot
  action recognition. In: International Joint Conference on Neural Networks
  (IJCNN). pp.~1--7 (2019)

\bibitem{f4Matching}
Vinyals, O., Blundell, C., Lillicrap, T., Kavukcuoglu, K., Wierstra, D.:
  Matching networks for one shot learning. In: Lee, D.D., Sugiyama, M., von
  Luxburg, U., Guyon, I., Garnett, R. (eds.) Advances in Neural Information
  Processing Systems 29: Annual Conference on Neural Information Processing
  Systems 2016, December 5-10, 2016, Barcelona, Spain. pp. 3630--3638 (2016)

\bibitem{lei_thesis_2017}
Wang, L.: Analysis and Evaluation of {K}inect-based Action Recognition
  Algorithms. Master's thesis, School of the Computer Science and Software
  Engineering, The University of Western Australia (Nov 2017)

\bibitem{lei_tip_2019}
Wang, L., Huynh, D.Q., Koniusz, P.: A comparative review of recent kinect-based
  action recognition algorithms. IEEE Transactions on Image Processing
  \textbf{29},  15--28 (2020)

\bibitem{lei_icip_2019}
Wang, L., Huynh, D.Q., Mansour, M.R.: Loss switching fusion with similarity
  search for video classification. ICIP  (2019)

\bibitem{10.1145/3474085.3475572}
Wang, L., Koniusz, P.: Self-Supervising Action Recognition by Statistical
  Moment and Subspace Descriptors, p. 4324–4333. Association for Computing
  Machinery, New York, NY, USA (2021),
  \url{https://doi.org/10.1145/3474085.3475572}

\bibitem{lei_iccv_2019}
Wang, L., Koniusz, P., Huynh, D.Q.: Hallucinating {IDT} descriptors and {I3D}
  optical flow features for action recognition with cnns. In: The IEEE
  International Conference on Computer Vision (ICCV) (October 2019)

\bibitem{stgcn2018aaai}
Yan, S., Xiong, Y., Lin, D.: {Spatial Temporal Graph Convolutional Networks for
  Skeleton-Based Action Recognition}. In: AAAI (2018)

\bibitem{pmlr-v139-yang21j}
Yang, C.H.H., Tsai, Y.Y., Chen, P.Y.: Voice2series: Reprogramming acoustic
  models for time series classification. In: Meila, M., Zhang, T. (eds.)
  Proceedings of the 38th International Conference on Machine Learning.
  Proceedings of Machine Learning Research, vol.~139, pp. 11808--11819. PMLR
  (18--24 Jul 2021)

\bibitem{hongguang2020eccv}
Zhang, H., Zhang, L., Qi, X., Li, H., Torr, P., Koniusz, P.: Few-shot action
  recognition with permutation-invariant attention. In: European Conference on
  Computer Vision (ECCV) (2020)

\bibitem{hao2021iclr}
Zhu, H., Koniusz, P.: Simple spectral graph convolution. In: International
  Conference on Learning Representations (ICLR) (2021)

\end{thebibliography}


\begin{thebibliography}{10}
\providecommand{\url}[1]{\texttt{#1}}
\providecommand{\urlprefix}{URL }
\providecommand{\doi}[1]{https://doi.org/#1}

\bibitem{bergstra2015hyperopt}
Bergstra, J., Komer, B., Eliasmith, C., Yamins, D., Cox, D.D.: Hyperopt: a
  python library for model selection and hyperparameter optimization.
  Computational Science \& Discovery  \textbf{8}(1),  014008 (2015),
  \url{http://stacks.iop.org/1749-4699/8/i=1/a=014008}

\bibitem{pmlr-v130-blondel21a_sup}
Blondel, M., Mensch, A., Vert, J.P.: Differentiable divergences between time
  series. In: Banerjee, A., Fukumizu, K. (eds.) Proceedings of The 24th
  International Conference on Artificial Intelligence and Statistics.
  Proceedings of Machine Learning Research, vol.~130, pp. 3853--3861. PMLR
  (13--15 Apr 2021)

\bibitem{marco2011icml_sup}
Cuturi, M.: Fast global alignment kernels. In: International Conference on
  Machine Learning (ICML) (2011)

\bibitem{marco2017icml_sup}
Cuturi, M., Blondel, M.: Soft-dtw: a differentiable loss function for
  time-series. In: International Con- ference on Machine Learning (ICML) (2017)

\bibitem{UCRArchive_supp}
Dau, H.A., Keogh, E., Kamgar, K., Yeh, C.C.M., Zhu, Y., Gharghabi, S.,
  Ratanamahatana, C.A., Yanping, Hu, B., Begum, N., Bagnall, A., Mueen, A.,
  Batista, G.: {The UCR Time Series Classification Archive} (October 2018),
  \url{https://www.cs.ucr.edu/~eamonn/time\_series\_data_2018/}

\bibitem{Li2010}
Li, W., Zhang, Z., Liu, Z.: {Action Recognition Based on A Bag of 3D Points}.
  In: CVPR. pp. 9--14 (2010)

\bibitem{Liu_2019_NTURGBD120_sup}
Liu, J., Shahroudy, A., Perez, M., Wang, G., Duan, L.Y., Kot, A.C.: Ntu rgb+d
  120: A large-scale benchmark for 3d human activity understanding. IEEE
  Transactions on Pattern Analysis and Machine Intelligence  (2019).
  \doi{10.1109/TPAMI.2019.2916873}

\bibitem{Oreifej2013}
Oreifej, O., Liu, Z.: {HON4D: Histogram of Oriented 4D Normals for Activity
  Recognition from Depth Sequences}. In: CVPR. pp. 716--723 (2013)

\bibitem{coltrane_sup}
Prabowo, A., Koniusz, P., Shao, W., Salim, F.D.: {COLTRANE:} convolutional
  trajectory network for deep map inference. In: Proceedings of the 6th {ACM}
  International Conference on Systems for Energy-Efficient Buildings, Cities,
  and Transportation, BuildSys 2019, New York, NY, USA, November 13-14, 2019.
  pp. 21--30. {ACM} (2019). \doi{10.1145/3360322.3360853}

\bibitem{RahmaniHOPC2014}
Rahmani, H., Mahmood, A., Huynh, D.Q., Mian, A.: {HOPC: Histogram of Oriented
  Principal Components of 3D Pointclouds for Action Recognition}. In: ECCV. pp.
  742--757 (2014)

\bibitem{su2022temporal_sup}
Su, B., Wen, J.R.: Temporal alignment prediction for supervised representation
  learning and few-shot sequence classification. In: International Conference
  on Learning Representations (2022)

\bibitem{uai_ke_sup}
Sun, K., Koniusz, P., Wang, Z.: Fisher-bures adversary graph convolutional
  networks. Conference on Uncertainty in Artificial Intelligence  \textbf{115},
   465--475 (2019)

\bibitem{action_domain_sup}
Tas, Y., Koniusz, P.: Cnn-based action recognition and supervised domain
  adaptation on 3d body skeletons via kernel feature maps. The British Machine
  Vision Conference (BMVC)  (2018)

\bibitem{www_sup}
Zhang, Y., Zhu, H., Meng, Z., Koniusz, P., King, I.: Graph-adaptive rectified
  linear unit for graph neural networks. In: Proceedings of the ACM Web
  Conference 2022. p. 1331–1339. WWW '22, Association for Computing
  Machinery, New York, NY, USA (2022). \doi{10.1145/3485447.3512159}

\bibitem{costa_sup}
Zhang, Y., Zhu, H., Song, Z., Koniusz, P., King, I.: Costa:
  Covariance-preserving feature augmentation for graph contrastive learning.
  ACM SIGKDD Conference on Knowledge Discovery and Data Mining (KDD)  (2022),
  \url{https://doi.org/10.1145/3534678.3539425}

\bibitem{zhu2021contrastive_sup}
Zhu, H., Sun, K., Koniusz, P.: Contrastive laplacian eigenmaps. Advances in
  Neural Information Processing Systems  \textbf{34} (2021)

\end{thebibliography}


\appendix

\pagestyle{headings}
\def\ECCVSubNumber{5350}  

\title{Uncertainty-DTW for Time Series and Sequences (Supplementary Material)}

\titlerunning{Uncertainty-DTW for Time Series and Sequences}
%
\author{Lei Wang$^{\star, \dagger, \S}$\orcidlink{0000-0002-8600-7099} \and
Piotr Koniusz$^{\star,\S,\dagger}$\orcidlink{0000-0002-6340-5289}}
\authorrunning{Wang and Koniusz}
%
\institute{$^{\dagger}$Australian National University \;
   $^\S$Data61/CSIRO\\
   $^\S$firstname.lastname@data61.csiro.au 
}

\maketitle

\thispagestyle{empty}

\setcounter{table}{5}
\setcounter{equation}{22}
\setcounter{figure}{6}

Below we provide additional analyses, protocols and details of our work.

\section{Datasets and their statistics}

\begin{table*}[!htbp]
\caption{UCR archive (the latest version from 2018) which we use for time series analysis. The information is grouped based on the type of time series.}
\begin{center}
\begin{tabular}{ l  c c c c}
\toprule
Dataset type & Avg. \#train & Avg. \#test & Total \#classes & Avg. length \\ 
\midrule
Device & 1261 & 1135 & 44 & 895\\
ECG & 708 & 1755 & 95 & 326\\
EOG & 362 & 362 & 24 & 1250\\
EPG & 40 & 249 & 6 & 601\\
Hemodynamics & 104 & 208 & 156 & 2000\\
HRM & 18 & 186 & 18 & 201 \\
Image & 595 & 1157 & 334 & 360\\
Motion & 347 & 1057 & 99 & 517\\
Power & 180 & 180 & 2 & 144 \\
Sensor & 420 & 1286 & 177 & 410\\
Simulated & 203 & 1021 & 32 & 267\\
Spectro & 179 & 147 & 24 & 553 \\
Spectrum & 305 & 388 & 17 & 1836\\
Traffic & 607 & 1391 & 12 & 24\\
Trajectory & 208 & 130 & 78 & 360\\
\bottomrule
\end{tabular}
\label{timeseries_dataset}
\end{center}
\end{table*}

\begin{table*}[!htbp]
\caption{Popular benchmark datasets which we use for few-shot action recognition.}
\begin{center}
\begin{tabular}{ l  c c c c c c c c }
\toprule
 Datasets & Year & Classes & Subjects & \#views & \#clips & Sensor & \#joints \\ 
\midrule
MSR Action 3D & 2010 & 20 & 10 & 1 & 567 & Kinect v1 & 20\\
3D Action Pairs & 2013 & 12 & 10 & 1 & 360 & Kinect v1 & 20\\
UWA 3D Activity & 2014 & 30 & 10 & 1 & 701 & Kinect v1 & 15\\
NTU RGB+D~& 2016 & 60 & 40 & 80 & 56,880 & Kinect v2 & 25\\
NTU RGB+D 120~& 2019 & 120 & 106 & 155 & 114,480 & Kinect v2 & 25\\
Kinetics-skeleton~& 2018 & 400 & - & - & $\sim$ 300,000 & - & 18 \\
\bottomrule
\end{tabular}
\label{datasets}
\end{center}
\end{table*}

\noindent{\bf The UCR time series archive  \citelatex{UCRArchive_supp}.} UCR, introduced in 2002, is an important resource in the time series analysis community with at least 1,000 published papers making use of at least 1 dataset from this archive. We use 128 datasets from the latest version of UCR from 2018, encompassing a wide variety of fields and lengths. Table \ref{timeseries_dataset} details the statistics of this archive by grouping the whole dataset into different types.

\noindent{\bf Few-shot action recognition datasets.} Table \ref{datasets} contains statistics of datasets used in our experiments. Smaller datasets listed below are used for more evaluations of supervised and unsupervised few-shot action recognition: 

\begin{itemize}
\item {\em{MSR Action 3D}}~\citelatex{Li2010} is an older AR dataset captured with the Kinect depth camera. It contains 20 human sport-related activities such as {\it jogging}, {\it golf swing} and {\it side boxing}. 

\item {{\em 3D Action Pairs}}~\citelatex{Oreifej2013} contains 6 selected pairs of actions that have very similar motion trajectories, \eg, {\it put on a hat} and {\it take off a hat}, {\it pick up a box} and {\it put down a box}, \etc. 

\item {{\em UWA 3D Activity}}~\citelatex{RahmaniHOPC2014} has 30 actions performed by 10 people of various height at different speeds in cluttered scenes. 
\end{itemize}

As MSR Action 3D, 3D Action Pairs, and UWA 3D Activity have not been used in FSAR, we created 10 training/testing splits, by choosing half of class concepts for training, and half for testing per split per dataset. Training splits were further subdivided for crossvalidation. Section \ref{small_proto} details the class concepts per split for small datasets.

\section{Table of notations}

Table~\ref{notations} (next page) shows the notations used in this paper with their short descriptions.

\newpage

\begin{table*}[!htbp]
\caption{Notations and their descriptions.}
\begin{center}
\begin{tabular}{c l}
\toprule
\rowcolor{gray!10}
 Notation &  Description\\
 \midrule
$\mPsi$ & Query feature maps\\
$\mPsi'$ &  Support feature maps\\
$\mPi$ & Path matrix\\
$\mD(\cdot,\cdot) $ & Pair-wise distances\\
$d^2_{*}(\cdot,\cdot)$ & Distance functions and * can be base (squared Euclidean), DTW, sDTW or uDTW\\
$\gamma$ & The relaxation parameter of sDTW/uDTW\\
$\tau$& The number of temporal blocks for query\\
$\tau'$& The number of temporal blocks for support\\
$\cov$& Pair-wise variances between all possible pairs of two sequences\\
$\cov^{\dag}$ & Element-wise inverse of $\cov$\\
$f(\cdot;\cdot)$ & Encoder function\\
$\mathcal{P}$ & The set of parameters to learn\\
$\beta$ & Regularization parameter\\
$\sigma$ & Uncertainty parameter\\
$\mX$ & Query frames per block\\
$\mX'$ & Support frames per block\\
$K$ & The size of dictionary\\
$K'$ & The subset size for $K'$ nearest anchors\\
$\vx$ & Time series for training\\
$\vx'$ & Time series for testing\\
$\vmu_c$ & Class prototype for class $c$\\
$\Omega(\cdot)$ & Regularization penalty\\
$\valpha$ & Coding vector\\
$\lambda_{\text{DL}}$ & Learning rate for dictionary learning\\
$\lambda_{\text{EN}}$ & Learning rate for encoder\\
$\mM$ & Dictionary anchors\\
$B$ & The number of training episodes\\
$N$ & The number of classes\\
$Z$ & The number of samples from each class\\
$J$& The number of human body joints\\
$d$ & Feature dimension after MLP\\
$d'$ & Feature dimension (output of EN)\\
$\delta$ & The similarity label\\
$N_c$ & The number of samples for class $c$ \\
\bottomrule
\end{tabular}
\label{notations}
\end{center}
\end{table*}

\newpage

\section{Evaluation Protocols}
\label{app:epr}

Below, we detail our new/additional evaluation protocols used in the experiments on few-shot action recognition.
\subsection{Few-shot AR protocols (the small-scale datasets)}
\label{small_proto}
As we use several  class-wise splits for small datasets, these splits will be simply released in our code. Below, we explain the selection process that we used.

\paragraph{FSAR (MSR Action 3D)}. As this dataset contains 20 action classes, we randomly choose 10 action classes for training and the rest 10 for testing. We repeat this sampling process 10 times to form in total 10 train/test splits. For each split, we have 5-way and 10-way experimental settings. The overall performance on this dataset is computed by averaging the performance over the 10 splits.

\paragraph{FSAR (3D Action Pairs)}. This dataset has in total 6 action pairs (12 action classes), each pair of action has very similar motion trajectories, \eg, {\it pick up a box} and {\it put down a box}. We randomly choose 3 action pairs to form a training set (6 action classes) and the half action pairs for the test set, and in total there are ${\binom nk}\!=\!{\binom {6}{3}\!=\!20}$ different combinations of train/test splits. As our train/test splits are based on action pairs, we are able to test  whether the algorithm is able to classify unseen action pairs that share similar motion trajectories. We use 5-way protocol on this dataset to evaluate the performance of FSAR, averaged over all 20 splits.

\paragraph{FSAR (UWA 3D Activity)}. This dataset has 30 action classes. We randomly choose 15 action classes for training and the rest half action classes for testing. We form in total 10 train/test splits, and we use 5-way and 10-way protocols on this dataset, averaged over all 10  splits.

\subsection{One-shot protocol on NTU-60}

Following NTU-120~\citelatex{Liu_2019_NTURGBD120_sup}, we introduce the one-shot AR setting on NTU-60. We split the whole dataset into two parts: auxiliary set (on NTU-120 the training set is called as auxiliary set, so we follow such a terminology) and one-shot evaluation set. 

\paragraph{Auxiliary set} contains 50 classes, and all samples of these classes can be used for learning and validation. Evaluation set consists of 10 novel classes, and one sample from each novel class is picked as the exemplar (terminology introduced by authors of NTU-120), while all the remaining samples of these classes are used to test the recognition performance. 

\paragraph{Evaluation set} contains 10 novel classes, namely A1, A7, A13, A19, A25, A31, A37, A43, A49, A55. 

The following 10 samples are the exemplars:

\noindent(01)S001C003P008R001A001, (02)S001C003P008R001A007,\\ (03)S001C003P008R001A013, (04)S001C003P008R001A019, \\(05)S001C003P008R001A025, (06)S001C003P008R001A031, \\(07)S001C003P008R001A037, (08)S001C003P008R001A043,\\ (09)S001C003P008R001A049, (10)S001C003P008R001A055. 


\paragraph{Auxiliary set} contains  50 classes (the remaining 50 classes of NTU-60 excluding the 10 classes in evaluation set).

\section{Effectiveness of SigmaNet}
\label{supp:barycenter}


In this section, we introduce several variants of how $\cov$ is computed to verify the effectiveness of our proposed SigmaNet.


Firstly, we investigate whether SigmaNet is needed in its current form (as in taking features to produce the uncertainty variable), or if $\cov$ could be learnt as the so-called free variable. To this end, 
we create a vector of parameters of size $\tau_{(0)}\!\cdot\!\tau_{(0)}$ which we register as one of parameters of the network (we backpropagate \wrt this parameter among others). We set  $\tau_{(0)}$ to be the average integer of numbers of blocks over sequences. 
We then reshape this vector into $\tau_{(0)}\!\times\!\tau_{(0)}$ matrix and initialize  with $0\!\pm\!0.1$ uniform noise. We then apply a 2D bilinear interpolation to the matrix to obtain $\cov$ of desired size $\tau\!\times\!\tau'$, where $\tau$ and $\tau'$ are the number of temporal blocks for query and support samples, respectively. The $\tau\!\times\tau'$ matrix is then passed into the sigmoid function to produce the $\cov$ matrix.

For classification of time series, 
we create a vector of parameters of size $t_{(0)}$  which we register as one of parameters of the network (we backpropagate \wrt this parameter among others). We set  $\tau_{(0)}$ to be the average integer of numbers of time steps of input time series. We initialize that vector with $0\!\pm\!0.1$ uniform noise, and we then use a 1D bilinear interpolation to interpolate the vector into desired length $\tau$. The interpolated vector is passed into the sigmoid function to generate $\boldsymbol{\sigma}_{\vx}$ for the input sequence $\vx$ of length $\tau$. For sequence $\vx'$ (exhaustive search via nearest neighbor) or $\vmu_c$ (via nearest centroid), we use exactly the same process to generate $\boldsymbol{\sigma}_{\vx'}$ or $\boldsymbol{\sigma}_{\vmu_c}$ but of course  they have their own vector of length $\tau_{(0)}$ that we minimize over. We obtain  $\cov\!=\!\boldsymbol{\sigma}^2_{\vx}\boldsymbol{1}^\top\!\!+\!\boldsymbol{1}{\boldsymbol{\sigma}_{\vx'}^\top}^2$  (or $\cov\!=\!\boldsymbol{\sigma}^2_{\vx}\boldsymbol{1}^\top\!\!+\!\boldsymbol{1}{\boldsymbol{\sigma}_{\vmu_c}^\top}^{\!2}$ if we use the nearest centroid), where squaring is performed in the element-wise manner.

\begin{table}[!htbp]
\centering
\caption{Comparisons of two different ways of generating $\cov$ for few-shot action recognition. Evaluations on the NTU-60 dataset.}
\begin{tabular}{lccccc}
\toprule
\#classes & 10 & 20 & 30 & 40 & 50\\
\midrule
uDTW ($\cov$ via the free variable) & 54.1 & 56.5 & 61.0 & 64.1 & 68.0\\
\rowcolor{blue!10}
uDTW ($\cov$ via SigmaNet)   & 56.9 & 61.2 & 64.8 & 68.3 & 72.4\\
\bottomrule
\end{tabular}
\label{tab:ntu_sigmanet}
\end{table}

\begin{table}[!htbp]
\caption{Comparisons of two different ways of generating $\cov$ for classification of time series. Evaluations on the UCR archive. K denotes the number of nearest neighbors used by the $K$ nearest neighbors based classification.}
\label{tab:ucr_sigmanet}
\begin{center}
\begin{tabular}{lcccc}
\toprule
 & \multicolumn{3}{c}{\bf Nearest neighbor} &{\bf Nearest centroid}\\
 \cline{2-4}
 & $K = 1$ & $K = 3$ & $K = 5$ & \\
\midrule
uDTW ($\cov$ via the free variable) & 77.0 & 77.3 & 78.0 & 70.9\\
\rowcolor{blue!10}
uDTW ($\cov$ via SigmaNet) & 80.0 & 81.2 &83.3 & 72.2 \\
\bottomrule
\end{tabular}
\end{center}
\end{table}

In conclusion, the above steps facilitate the direct minimization \wrt the variable tied with $\cov$ instead of learning $\cov$ through our SigmaNet whose input are encoded features \etc. Tables \ref{tab:ntu_sigmanet} and \ref{tab:ucr_sigmanet} show that using SigmaNet is a much better choice than trying to infer the uncertainty by directly minimizing the free variable. The result is expected as SigmaNet learns to associate feature patterns of sequences with their uncertainty patterns. Minimizing \wrt the free variables cannot learn per se.

\section{Additional Visualizations of Forecasting the Evolution of Time Series}
\label{supp:pred}

\begin{figure}[t]
\centering
\begin{subfigure}[t]{0.45\linewidth}
\includegraphics[trim=12.6cm 0.5cm 0.6cm 9cm, clip=true,width=0.99\linewidth]{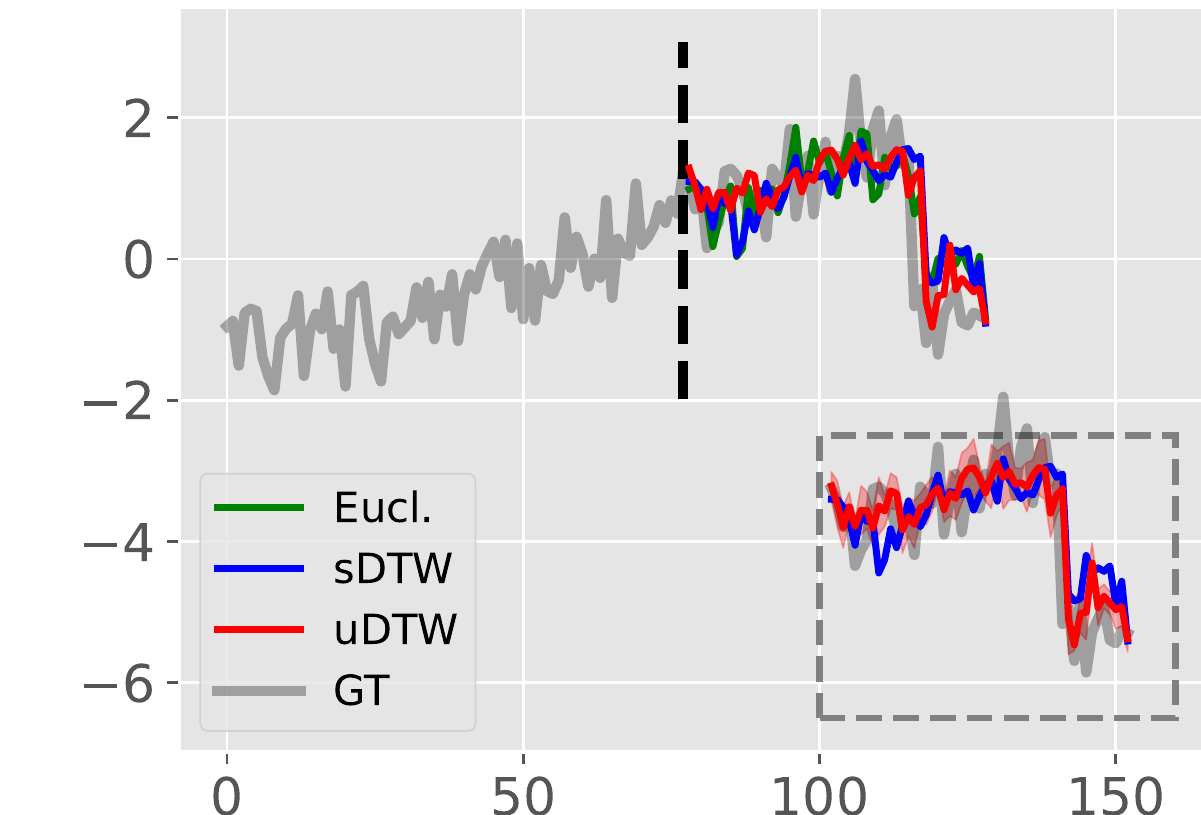}
\caption{CBF}\label{fig:cbf_more}
\end{subfigure}
\begin{subfigure}[t]{0.45\linewidth}
\includegraphics[trim=12.6cm 0.5cm 0.6cm 9cm, clip=true,width=0.99\linewidth]{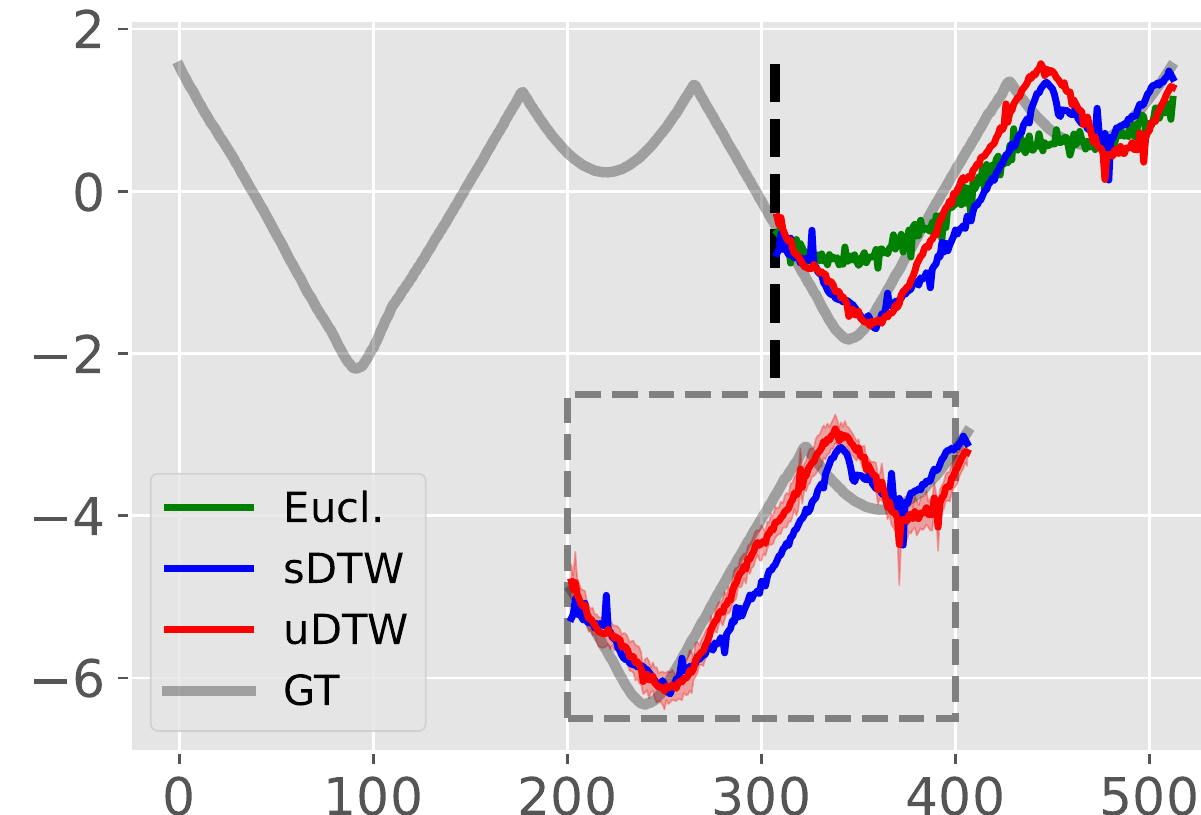}
\caption{ShapesAll}\label{fig:shapeall_more}
\end{subfigure}
\caption{Additional visualizations for forecasting the evolution of time series. Given the first part of a time series, we train 
the pipeline from Fig. \ref{fig:forecast} to predict the remaining part of the time series. We compare the use of the Euclidean, sDTW or uDTW distances within the pipeline. We use CBF and ShapesAll in UCR archive, and display the prediction obtained
for the given test sample with either of these 3 distances, and the ground truth (GT).
Oftentimes, we observe that uDTW helps predict the sudden changes well. (a) Our uDTW aligns well with the ground truth compared to sDTW. (b) Our uDTW generates better shape of prediction compared to sDTW (for example note the red curve following much closer the rising gray slope). Quantitative results of MSE and `shape' metrics for the whole UCR archive are given in the main paper.}
\label{fig:pred_more}
\end{figure}

We provide additional visualizations of forecasting the evolution of time series in Figure~\ref{fig:pred_more}. We notice that our uDTW produces predictions that are better aligned with the ground truth (see Fig.~\ref{fig:cbf_more}). Moreover, our uDTW generates better shape of the predictions compared to sDTW, and the predictions from sDTW have more perturbations/fluctuations (see Fig.~\ref{fig:shapeall_more}). Quantitative results for the whole UCR archive can be found in the main paper.



\section{Additional Evaluations for Few-shot Action Recognition}
\label{supp:fsl}

\begin{table}[!htbp]
\caption{uDTW derived under the Normal, Laplacian and Cauchy distributions. Evaluations of few-shot action recognition  on small-scale datasets.}
\label{tab:fsl-small}
\vskip 0.15in
\begin{center}
\begin{tabular}{lcccccc}
\toprule
& \multicolumn{3}{c}{\bf Supervised} & \multicolumn{3}{c}{\bf Unsupervised}\\
\cline{2-7}
& MSR  & $\;$3D Action Pairs & $\;$UWA 3D & $\;$MSR & $\;$3D Action Pairs & $\;$UWA 3D\\
\midrule
TAP (HM)~\citelatex{su2022temporal_sup} & 67.40 & 77.22 & 37.13 & - & - & - \\
TAP (Lifted)~\citelatex{su2022temporal_sup} & 65.20 & 78.33 & 34.80 & - & - & - \\
TAP (Bino.)~\citelatex{su2022temporal_sup} & 66.67 & 78.33 & 36.55 & - & - & - \\
sDTW~\citelatex{marco2017icml_sup} & 70.59 & 81.67 & 44.74 & 62.63 & 48.33 & 39.47 \\
\hdashline
uDTW (Laplace) & 72.24 & 82.89 & 45.64 & {\bf 66.00} & {\bf 55.00}  & 41.22 \\
uDTW (Cauchy) & 70.88 & {\bf84.44} & 45.03 & 65.12 &  50.32 & 40.50 \\
uDTW (Normal) & {\bf 72.66} & 83.33 & {\bf47.66} & 65.00 & 52.22  & {\bf 41.74} \\
\bottomrule
\end{tabular}
\end{center}
\vskip -0.1in
\end{table}

\begin{table}[!htbp]
\centering
\caption{uDTW derived under the Normal, Laplacian and Cauchy distributions.  Evaluations of few-shot action recognition on the large-scale NTU-60 dataset.}
\begin{tabular}{lccccc}
\toprule
\#classes & 10 & 20 & 30 & 40 & 50\\
\midrule
& \multicolumn{5}{c}{\bf Supervised}\\
\hline
sDTW(baseline)~\citelatex{marco2017icml_sup}& 53.7 & 56.2 & 60.0 & 63.9 & 67.8\\
\hdashline
uDTW(Cauchy)& 56.1 & 61.1 & 62.9 & 68.3 & 69.9\\
uDTW(Laplace)& 55.3 & 59.2 & 63.3 & 67.7 & 70.3\\
\rowcolor{blue!10}
uDTW(Normal)& 56.9 & 61.2 & 64.8 & 68.3 & 72.4 \\
\hline
& \multicolumn{5}{c}{\bf Unsupervised}\\
\hline
sDTW(baseline)~\citelatex{marco2017icml_sup} & 35.6 & 45.2 & 53.3 & 56.7 & 61.7 \\
\hdashline
uDTW(Cauchy) & 36.7& 47.9 & 54.9 & 57.3& 63.3\\
uDTW(Laplace) & 36.2 & 48.2 & 54.3 & 57.8& 63.1\\
\rowcolor{blue!10}
uDTW(Normal) & 37.0 & 48.3 & 55.3 & 58.0 & 63.3\\
\bottomrule
\end{tabular}
\label{tab:ntu_other_dist}
\end{table}

We also evaluate our proposed uDTW versus sDTW on smaller datasets for both supervised and unsupervised settings. As uDTW was derived in Section \ref{sec:der} under modeling the MLE of the product of the Normal distributions, we investigate modeling each path $\mPi_i$ by replacing the Normal distribution with the Laplace or Cauchy distributions. By applying MLE principles in analogy to Section \ref{sec:der}, we arrive at $\beta\Omega_{\mPi_i} +d^2_{\mPi_i}$ for 
%
%
\begin{enumerate}
    \item Laplace: $\sum_{(m,n)\in\mPi_i} \beta\log(\sigma_{mn}) +\frac{\|\vpsi_m\!-\!\vpsi'_n\|_1}{\sigma_{mn}}$;
    \item Cauchy: $\sum_{(m,n)\in\mPi_i} \beta\log(\sigma_{mn}) +\log \left(1 + \frac{\|\vpsi_m\!-\!\vpsi'_n\|_2^2}{\sigma^2_{mn}}\right)$.
\end{enumerate}

Table~\ref{tab:fsl-small} shows that uDTW  achieves better performance than sDTW, and the Laplace distribution is performing particularly well on the unsupervised few-shot action recognition.  
Table~\ref{tab:ntu_other_dist} shows that uDTW based on the Normal distribution is overall better than other distributions on large-scale datasets such as NTU-60. For this very reason we use uDTW based on the Normal distribution.

\section{Additional Visualizations on Barycenters}

\begin{figure}[!htbp]
\centering
\begin{subfigure}[t]{0.9\linewidth}
\centering\includegraphics[trim=4cm 1.2cm 3.5cm 0.6cm, clip=true,width=\linewidth]{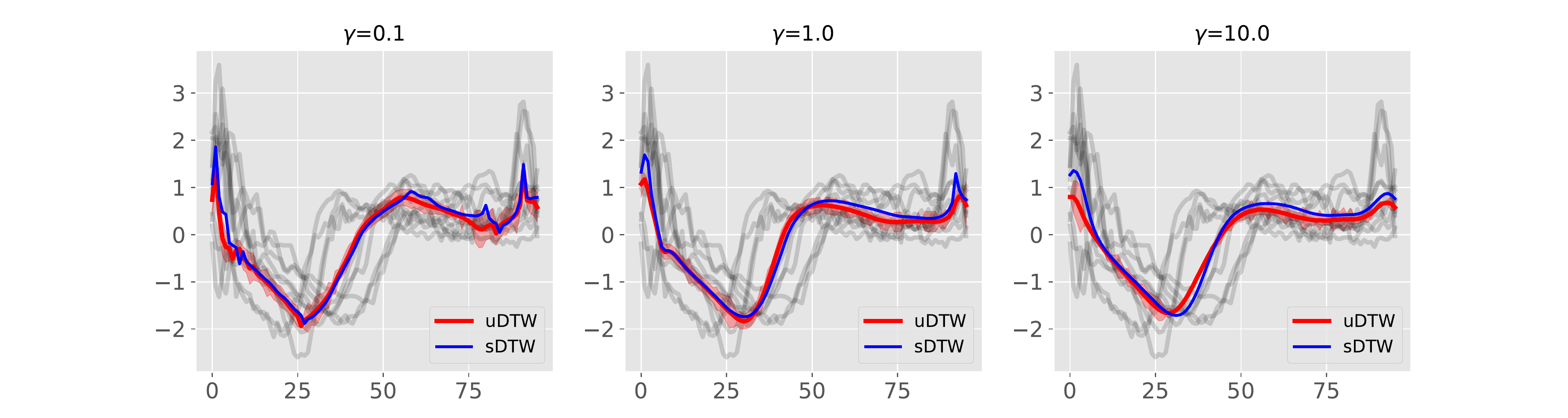}
\caption{ECG200}\label{fig:ECG200}
\end{subfigure}
\begin{subfigure}[t]{0.9\linewidth}
\centering\includegraphics[trim=4cm 1.2cm 3.5cm 0.6cm, clip=true,width=\linewidth]{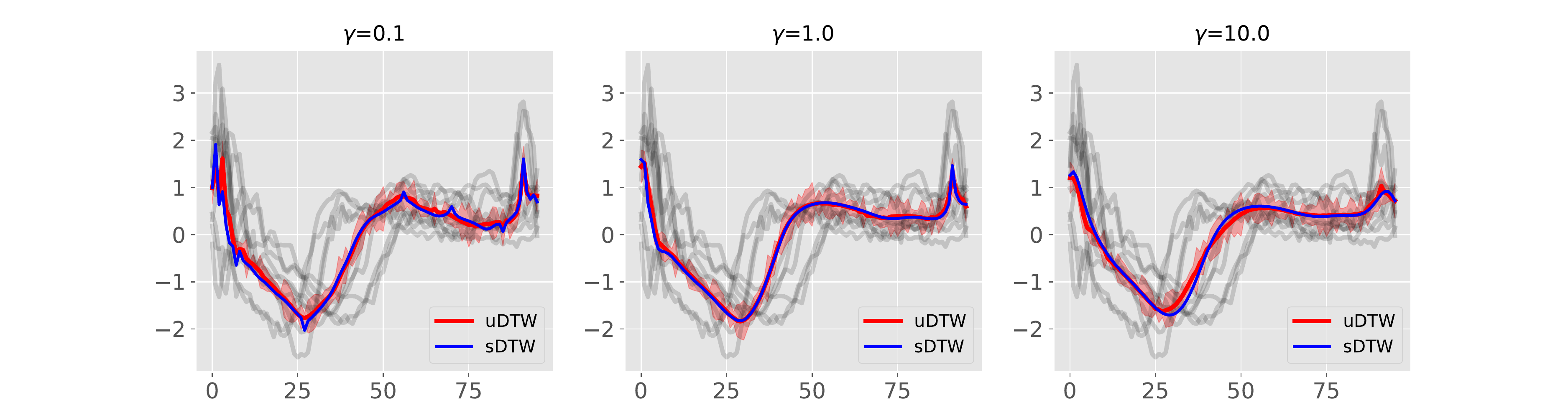}
\caption{ECG200 (our SigmaNet)}\label{fig:ECG200_sn}
\end{subfigure}
\begin{subfigure}[t]{0.9\linewidth}
\centering\includegraphics[trim=4cm 1.2cm 3.5cm 0.6cm, clip=true,width=\linewidth]{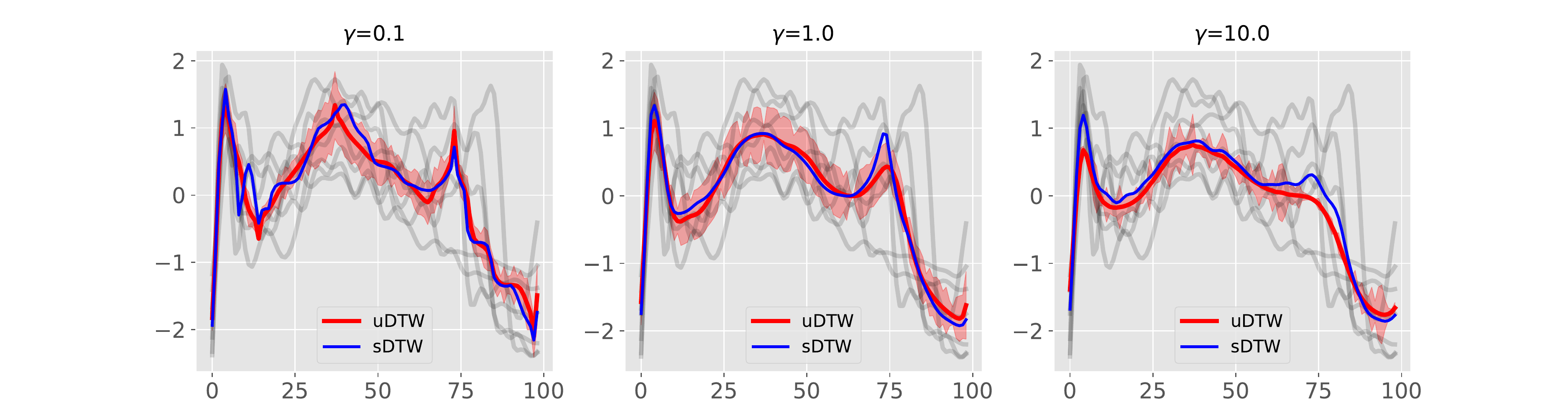}
\caption{Medical Images}\label{fig:mi}
\end{subfigure}
\begin{subfigure}[t]{0.9\linewidth}
\centering\includegraphics[trim=4cm 1.2cm 3.5cm 0.6cm, clip=true,width=\linewidth]{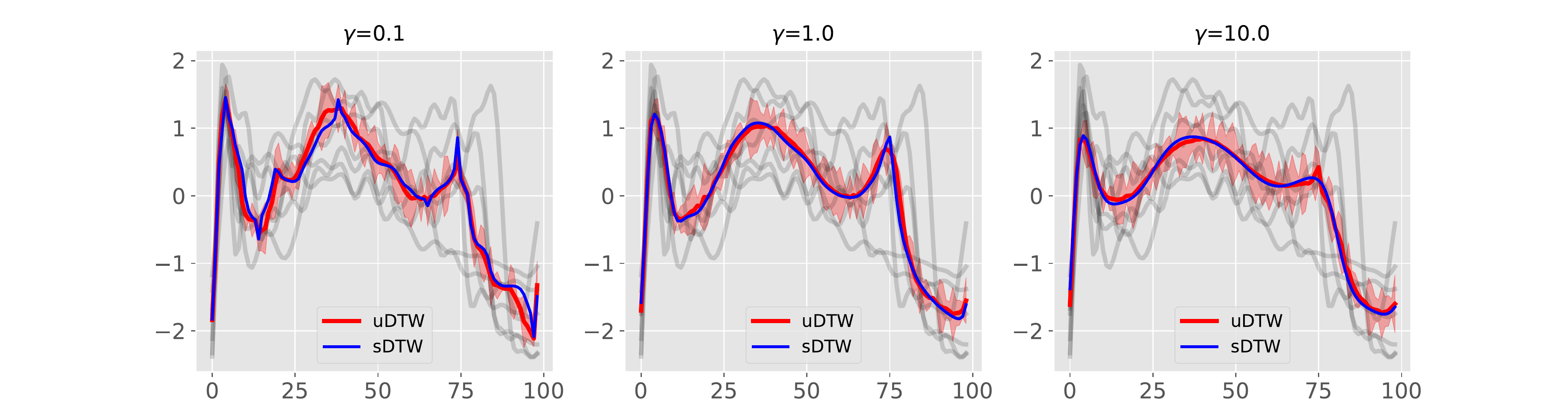}
\caption{Medical Images (our SigmaNet)}\label{fig:mi_sn}
\end{subfigure}
\caption{Comparison of barycenters based on our uDTW \vs sDTW.
We visualize uncertainty around the barycenters in red color for uDTW. Our uDTW with SigmaNet generates reasonable barycenters even when higher $\gamma$ values are used, \eg, $\gamma\!=\!10.0$. Higher $\gamma$ value leads to smooth barycenters but introducing higher uncertainty.}
\label{fig:wwosigmanet}
\end{figure}

Figure~\ref{fig:wwosigmanet} shows more visualizations of barycenters of time series. With our SigmaNet, we  obtain much better barycenters with our uDTW compared to sDTW. 

\newpage

\section{Network Configuration and Training Details}
\label{network_train}

Below we provide the details of network configuration and training process. 



\subsection{Skeleton Data Preprocessing}

Before passing the skeleton sequences into MLP and a simple linear graph network (\eg, S$^2$GC), we first normalize each body joint \wrt to the torso joint ${\bf v}_{f, c}$:
\begin{equation}
    {\bf v}^\prime_{f, i}\!=\!{\bf v}_{f, i}\!-\!{\bf v}_{f, c},
\end{equation}
where $f$ and $i$ are the index of video frame and human body joint respectively. After that, we further normalize each joint coordinate into  [-1, 1] range:

\begin{equation}
    \hat{{\bf v}}_{f, i}[j] = \frac{{\bf v}^\prime_{f, i}[j]}{ \text{max}([\text{abs}({\bf v}^\prime_{f, i}[j])]_{f\in{\mathcal{I}_\tau},i\in\mathcal{I}_{J} } )},
\end{equation}
where $j$ is for selection of the $x$, $y$ and $z$ axes, $\tau$ is the number of frames and $J$ is the number of 3D body joints per frame.

For the skeleton sequences that have more than one performing subject, (i) we normalize each skeleton separately, and each skeleton is passed to MLP for learning the temporal dynamics, and (ii) for the output features per skeleton from MLP, we pass them separately to the graph neural network, \eg, two skeletons from a given video sequence will have two outputs obtained from the graph neural network, and we aggregate the outputs through average pooling before passing to sDTW or uDTW. 

\subsection{Network Configuration}
\noindent{\bf SigmaNet.} It is composed of an FC layer and a scaled sigmoid function which translate the learned features of either actions or time series into desired $\cov$. 
The input to FC is of the size of feature dimension (depends on the encoder) and the output is a scalar. SigmaNet with the scaled sigmoid function can be defined as: 
\begin{equation}
    \sigma(\vpsi)\!=\!\frac{\kappa}{1\!+\!\exp({-\text{FC}(\vpsi)})}\!+\!\eta,
    \label{eq:sigmoid}
\end{equation}
where $\eta\!>\!0$ is the offset and $\kappa\!\geq\!0$ is the maximum magnitude of sigmoid. For an entire sequence with $\tau$ blocks, the SigmaNet produces vector $\boldsymbol{\sigma}_\vx$ for sequence $\vx$ and $\boldsymbol{\sigma}_{\vx'}$ for sequence $\vx'$ (we concatenate per-block scalars to form these vectors), and we typically obtain   $\cov\!=\!\boldsymbol{\sigma}^2_\vx\textbf{1}^\top\!+\!{\mathbf{1}\boldsymbol{\sigma}^2_{\vx'}}^{\!\top}$.

\paragraph{Forecasting of the evolution of time series.} The MLP for this task consists of two FC layers with a tanh layer in between. The input to the first FC layer is $t$ and output size is $t'$, and after the tanh layer, the input to the second FC layer is $t'$ and output $(\tau\!-\!t)$ dimensional prediction. We set $t'\!=30$ or 50 depending on the length of time series in each dataset.

\paragraph{Few-shot action recognition.} Given the temporal block size $M$ (the number of frames in a block) and desired output size $d$, the configuration of the 3-layer MLP unit is: FC ($3M \rightarrow 6M$), LayerNorm (LN) as in \cite{dosovitskiy2020image}, ReLU, FC ($6M \rightarrow 9M$), LN, ReLU, Dropout (for smaller datasets, the dropout rate is 0.5; for large-scale datasets, the dropout rate is 0.1), FC ($9M \rightarrow d$), LN. Note that $M$ is the temporal block size
~and $d$ is the output feature dimension per body joint. We set $M\!=\!10$ for experiments.

For the encoding network, let us take the query input $\mathbf{X}\!\in\!\mathbb{R}^{3\times J\times M}$ for the temporal block of length $M$ as an example, where $3$ indicates that Cartesian coordinates $(x,y,z)$ were used, and $J$ is the number of body joints. As alluded to earlier, we obtain $\widehat{\mathbf{X}}^T\!=\!\text{MLP}(\mathbf{X}; \mathcal{P}_{MLP})\!\in\!\mathbb{R}^{{d}\times J}$.

Subsequently, we employ a simple linear graph network, S\textsuperscript{2}GC from Section \ref{sec:ssgc},  and the transformer encoder \cite{dosovitskiy2020image} which consists of alternating layers of Multi-Head Self-Attention (MHSA) and a feed-forward MLP (two FC layers with a GELU non-linearity between them). LayerNorm (LN) is applied before every block, and residual connections after every block. Each block feature matrix $\widehat{\mathbf{X}} \in \mathbb{R}^{J \times {d}}$ encoded by a simple linear graph network S\textsuperscript{2}GC (without learnable $\bf \Theta$) is then passed to the transformer. Similarly to the standard transformer, we prepend a learnable vector  
${\bf y}_\text{token}\!\in\!\mathbb{R}^{1\times {d}}$ to the sequence of block features $\widehat{\mathbf{X}}$ obtained from S\textsuperscript{2}GC,
and we also add the positional embeddings ${\bf E}_\text{pos} \in \mathbb{R}^{(1+J) \times {d}}$ based on the sine and cosine functions (standard in transformers) so that token ${\bf y}_\text{token}$ and each body joint enjoy their own unique positional encoding. 
We obtain  $\mathbf{Z}_0\!\in\!\mathbb{R}^{(1+J)\times {d}}$ which is the input in the following backbone:
\begin{align}
&{\bf Z}_0 = [{\bf y}_\text{token}; \text{S\textsuperscript{2}GC}(\widehat{\mathbf{X}})]+{\bf E}_\text{pos}, \label{eq:proj}\\
&{\bf Z}^\prime_k = \text{MHSA}(\text{LN}({\bf Z}_{k-1})) + {\bf Z}_{k-1}, \;k = 1, \cdots, L_\text{tr}\label{eq:mhsa}\\
& {\bf Z}_k = \text{MLP}(\text{LN}({\bf Z}^\prime_k)) + {\bf Z}^\prime_k, \qquad\quad \,k = 1, \cdots, L_\text{tr}\label{eq:mlp} \\
& \mathbf{y}' = \text{LN}\big(\mathbf{Z}^{(0)}_{L_\text{tr}}\big) \qquad\qquad\text{ where }\quad\;\; \mathbf{y}'\in \mathbb{R}^{1 \times {d}} \label{eq:blockfeat}\\
&f(\mathbf{X}; \mathcal{P})=\text{FC}(\mathbf{y}'^T; \mathcal{P}_{FC})\qquad\qquad\quad\!\in \mathbb{R}^{d'},\label{eq:final_fc_bl}
\end{align}
 \sloppy where $\mathbf{Z}^{(0)}_{L_\text{tr}}$ is the first ${d}$ dimensional row vector extracted from the output matrix $\mathbf{Z}_{L_\text{tr}}$  of size $(J\!+\!1)\times{d}$ which corresponds to the last layer $L_\text{tr}$ of the transformer. Moreover, parameter $L_\text{tr}$ controls the depth of the transformer, whereas $\mathcal{P}\!\equiv\![\mathcal{P}_{MLP},\mathcal{P}_{S\textsuperscript{2}GC},\mathcal{P}_{Transf},\mathcal{P}_{FC}]$ is the  set of  parameters of EN. In case of  S$^2$GC, $|\mathcal{P}_{S\textsuperscript{2}GC}|\!=\!0$ because we do not use their learnable parameters $\bf \Theta$ (\ie, think $\bf \Theta$ is set as the identity matrix in Eq. \eqref{eq:ssgc}).

We can define now a support feature map as $\vec{\Psi}'\!=\![f(\boldsymbol{X}_1;\mathcal{P}),\cdots,f(\boldsymbol{X}_{\tau'};\mathcal{P})]\!\in\!\mathbb{R}^{d'\times\tau'}$ for $\tau'$ temporal blocks, and the query map $\vec{\Psi}$ accordingly.

 The hidden size of our transformer  (the output size of the first FC layer of the MLP depends on the dataset. For smaller datasets, the depth of the transformer is $L_\text{tr}\!=\!6$ with $64$ as the hidden size, and the MLP output size is  ${d}\!=\!32$ (note that the MLP which provides $\widehat{\mathbf{X}}$ and the MLP in the transformer must both have  the same output size). For NTU-60, the depth of the transformer is $L_\text{tr}\!=\!6$, the hidden size is 128 and the MLP output size is  ${d}\!=\!64$. For NTU-120, the depth of the transformer is   $L_\text{tr}\!=\!6$, the hidden size is 256 and the MLP size is ${d}\!=\!128$. For Kinetics-skeleton, the depth for the transformer is $L_\text{tr}\!=\!12$, hidden size is 512 and the MLP output size is ${d}\!=\!256$. The number of Heads for the transformer of smaller datasets, NTU-60, NTU-120 and Kinetics-skeleton is set as 6, 12, 12 and 12, respectively.

The output sizes $d'$ of the final FC layer are 50, 100, 200, and 500 for the smaller datasets, NTU-60, NTU-120 and Kinetics-skeleton, respectively.

\subsection{Linear Graph Network (S\textsuperscript{2}GC)}
\label{sec:ssgc}
Based on a modified Markov Diffusion Kernel, Simple Spectral Graph Convolution (S\textsuperscript{2}GC) is the summation over $l$-hops, $l\!=\!1,\cdots,L$.  The output of S\textsuperscript{2}GC is given as:
\begin{equation}
    {\bf \Phi_\text{S\textsuperscript{2}GC}} \!=\!  \frac{1}{L}\sum_{l=1}^{L}((1\!-\!\alpha){\bf S}^l{\bf X}\!+\!\alpha{\bf X}){\bf \Theta},
    \label{eq:ssgc}
\end{equation}
where $L\!\geq\!1$ is the number of linear layers and $\alpha\!\geq\!0$ determines the importance of self-loop of each node (we use their default setting $\alpha\!=\!0.05$ and $L\!=\!6$). Choice of other graph embeddings are possible, including contrastive models COLES \citelatex{zhu2021contrastive_sup} or COSTA \citelatex{costa_sup}, adversarial Fisher-Bures GCN \citelatex{uai_ke_sup} or GCNs with rectifier attention  \citelatex{www_sup}. 
One may also use kernels on 3D body joints as in \citelatex{action_domain_sup} or even use CNN to encode 3D  body joints  as COLTRANE \citelatex{coltrane_sup}.

\subsection{$K$-NN classifier with SoftMax}
\label{sec:softmax}

For the $K$-NN classifier, instead of using $K$ best weights proportional to the inverse of the distance from the query sample $\vx^*$ to the closest samples $\vx_n$ (as is done in the soft-DTW paper \citelatex{marco2017icml_sup}) and expressed by
\begin{equation}
w(\vx_n | \vx^*) = \frac{1}{d^2\big(\vx^*, \vx_n\big)},
\label{eq:knn_basic}
\end{equation}
we weigh the neighbors $\vx_n$ of $\vx^*$ using 
\begin{equation}
%
w(\vx_n | \vx^*) = \frac{\text{exp}\big(-\frac{1}{\gamma''}d^2\big(\vx^*, \vx_n\big)\big)}{\sum_{n'\in\mathcal{N}(\vx^*; K)}\text{exp}\big(-\frac{1}{\gamma''}d^2\big(\vx^*, \vx_{n'}\big)\big)}
\label{eq:knn_soft}
\end{equation}
such that $\mathcal{N}(\vx^*; K)$ produces $K$ nearest samples $\vx_{n'}$ of $\vx^*$ according to  distance $d(\cdot,\cdot)$, \eg, the Euclidean distance, sDTW or uDTW. 
Parameter $\gamma''\!>\!0$ (in our case, we set $\gamma''\!=\!6$) further controls the impact of each sample $\vx_{n}$ on the classifier based on the bell shape of Radial Basis Function in the above equation.

Table~\ref{tab:ucrclassification_knn} shows the comparisons. We notice that the use of SoftMax in the $K$-NN classifier improves the performance for all the methods when $K\!=\!3$ and $K\!=\!5$.

\begin{table}[t]
\caption{Classification accuracy (mean$\pm$std) on UCR archive using nearest neighbor. $K$ denotes the number of nearest neighbors in the $K$-NN classifier. Highlighted rows are the based on SoftMax from Eq. \eqref{eq:knn_soft}. Non-highlighted rows are based on Eq. \eqref{eq:knn_basic}.}
\label{tab:ucrclassification_knn}
\vskip 0.15in
\begin{center}
\begin{tabular}{lccc}
\toprule
 & \multicolumn{3}{c}{\bf Nearest neighbor} \\
 \cline{2-4}

 & $K = 1$ & $K = 3$ & $K = 5$\\
\midrule
Euclidean & 71.2$\pm$17.5& 69.5$\pm$18.0& 67.5$\pm$17.6\\
\rowcolor{blue!10}
Euclidean (SoftMax) & 71.2$\pm$17.5& 72.3$\pm$18.1& 73.0$\pm$16.7\\
DTW~\citelatex{marco2011icml_sup} & 74.2$\pm$16.6&72.8$\pm$16.9&71.4$\pm$16.8\\
\rowcolor{blue!10}
DTW~\citelatex{marco2011icml_sup} (SoftMax) & 74.2$\pm$16.6&75.0$\pm$17.0&75.4$\pm$15.8\\
sDTW~\citelatex{marco2017icml_sup} & 76.2$\pm$16.6& 74.0$\pm$15.6&70.5$\pm$17.6\\
\rowcolor{blue!10}
sDTW~\citelatex{marco2017icml_sup} (SoftMax) & 76.2$\pm$16.6&77.2$\pm$15.9& 78.0$\pm$16.5\\
sDTW div.~\citelatex{pmlr-v130-blondel21a_sup} & 78.6$\pm$16.2& 76.5$\pm$16.4&74.8$\pm$15.8\\
\rowcolor{blue!10}
sDTW div.~\citelatex{pmlr-v130-blondel21a_sup} (SoftMax)& 78.6$\pm$16.2& 79.5$\pm$16.7&80.1$\pm$16.5\\
\hdashline
uDTW & 80.0$\pm$15.0& 78.0$\pm$15.8&76.2$\pm$15.0\\
\rowcolor{blue!10}
uDTW (SoftMax) & 80.0$\pm$15.0& 81.2$\pm$17.8&83.3$\pm$16.2\\

\bottomrule
\end{tabular}
\end{center}
\vskip -0.1in
\end{table}

\subsection{Training Details}
For both time series and few-shot action recognition pipelines, the weights are initialized with the normal distribution (zero mean and unit standard deviation).
We use 1e-3 for the learning rate, and the weight decay is 1e-6. We use the SGD optimizer. 

For time series, we set the training epochs to 30, 50 and 100 depending on the dataset in the UCR archive (due to many datasets, the epoch settings will be provided in the code directly). 

For few-shot action recognition, we set the number of training episodes to 100K for NTU-60, 200K for NTU-120, 500K for 3D Kinetics-skeleton, 10K for small datasets such as UWA 3D Multiview Activity II.

\section{Hyperparameters Evaluation}

In this section, we evaluate the impact of key hyperparameters. Remaining hyperparameters are obtained through Hyperopt \citelatex{bergstra2015hyperopt} for hyperparameter search on the validation set. 

\subsection{Evaluation of $\cov$}

We compare results given different formulations of $\cov$ in Tables~\ref{tab:smallerset_sigma} and \ref{tab:ntu60_sigma}. We notice that on smaller datasets, it is hard to determine which variant of $\cov$  is better (as these earlier datasets have fewer limited reliable skeletons compared to the new datasets). However, on bigger datasets, $\cov\!=\!\boldsymbol{\sigma}^2_{\vpsi}\boldsymbol{1}^\top\!\!+\!\boldsymbol{1}{\boldsymbol{\sigma}_{\vpsi'}^\top}^{\!2}$ performs the best in all cases; thus we choose this formulation of  $\cov$  for large-scale datasets. 

\begin{table}[t]
\caption{Evaluation of different variants of $\cov$ computation on small-scale datasets (supervised few-shot action recognition). Operator $\odot$ is the Hadamart product.}
\label{tab:smallerset_sigma}
\centering
\begin{tabular}{lcccc}
\toprule
 & $\;\boldsymbol{\sigma}_{\vpsi}\boldsymbol{1}^\top\!\!\!\odot\!\boldsymbol{1}{\boldsymbol{\sigma}_{\vpsi'}^\top}\;$ & $\;\boldsymbol{\sigma}^2_{\vpsi}\boldsymbol{1}^\top\!\!\!\odot\!\boldsymbol{1}{\boldsymbol{\sigma}_{\vpsi'}^\top}^{\!2}\;$ & $\;\boldsymbol{\sigma}_{\vpsi}\boldsymbol{1}^\top\!\!\!+\!\boldsymbol{1}{\boldsymbol{\sigma}_{\vpsi'}^\top}\;$ & $\;\boldsymbol{\sigma}^2_{\vpsi}\boldsymbol{1}^\top\!\!\!+\!\boldsymbol{1}{\boldsymbol{\sigma}_{\vpsi'}^\top}^{\!2}$\\
\midrule
MSR Action 3D & {\bf 72.32} & 68.51 & 70.59 & 69.20\\
3D Action Pairs& 82.78 & 80.56 & 82.22 & {\bf 85.00} \\
UWA 3D Activity & 43.86 & {\bf 45.91} & {\bf 45.91} & 45.03 \\
\bottomrule
\end{tabular}
\end{table}


\begin{table}[t]
\caption{Evaluation of different variants of $\cov$ computation on the large-scale NTU-60 dataset (supervised few-shot action recognition). }
\label{tab:ntu60_sigma}
\centering
\begin{tabular}{lcccc}
\toprule
\#classes$\;\;$ & $\;\boldsymbol{\sigma}_{\vpsi}\boldsymbol{1}^\top\!\!\!\odot\!\boldsymbol{1}{\boldsymbol{\sigma}_{\vpsi'}^\top}\;$ & $\;\boldsymbol{\sigma}^2_{\vpsi}\boldsymbol{1}^\top\!\!\!\odot\!\boldsymbol{1}{\boldsymbol{\sigma}_{\vpsi'}^\top}^{\!2}\;$ & $\;\boldsymbol{\sigma}_{\vpsi}\boldsymbol{1}^\top\!\!\!+\!\boldsymbol{1}{\boldsymbol{\sigma}_{\vpsi'}^\top}\;$ & $\;\boldsymbol{\sigma}^2_{\vpsi}\boldsymbol{1}^\top\!\!\!+\!\boldsymbol{1}{\boldsymbol{\sigma}_{\vpsi'}^\top}^{\!2}$\\
\midrule
$10$ & 56.6 & 56.0 & 55.6 & {\bf 56.9}\\
$20$ & 60.4 & 61.0 & 61.2 & {\bf 61.2} \\
$30$ & 64.2 & 64.1 & 63.5 & {\bf 64.8} \\
$40$ & 68.1 & 66.9 & 67.2 & {\bf 68.3} \\
$50$ & 72.0 & 72.3 & 72.0 & {\bf 72.4} \\
\bottomrule
\end{tabular}
\end{table}

\subsection{Evaluation of $\kappa$ and $\eta$ of SigmaNet}

\begin{figure}[!htbp]
\centering
\begin{subfigure}[t]{0.32\linewidth}
\includegraphics[width=\linewidth]{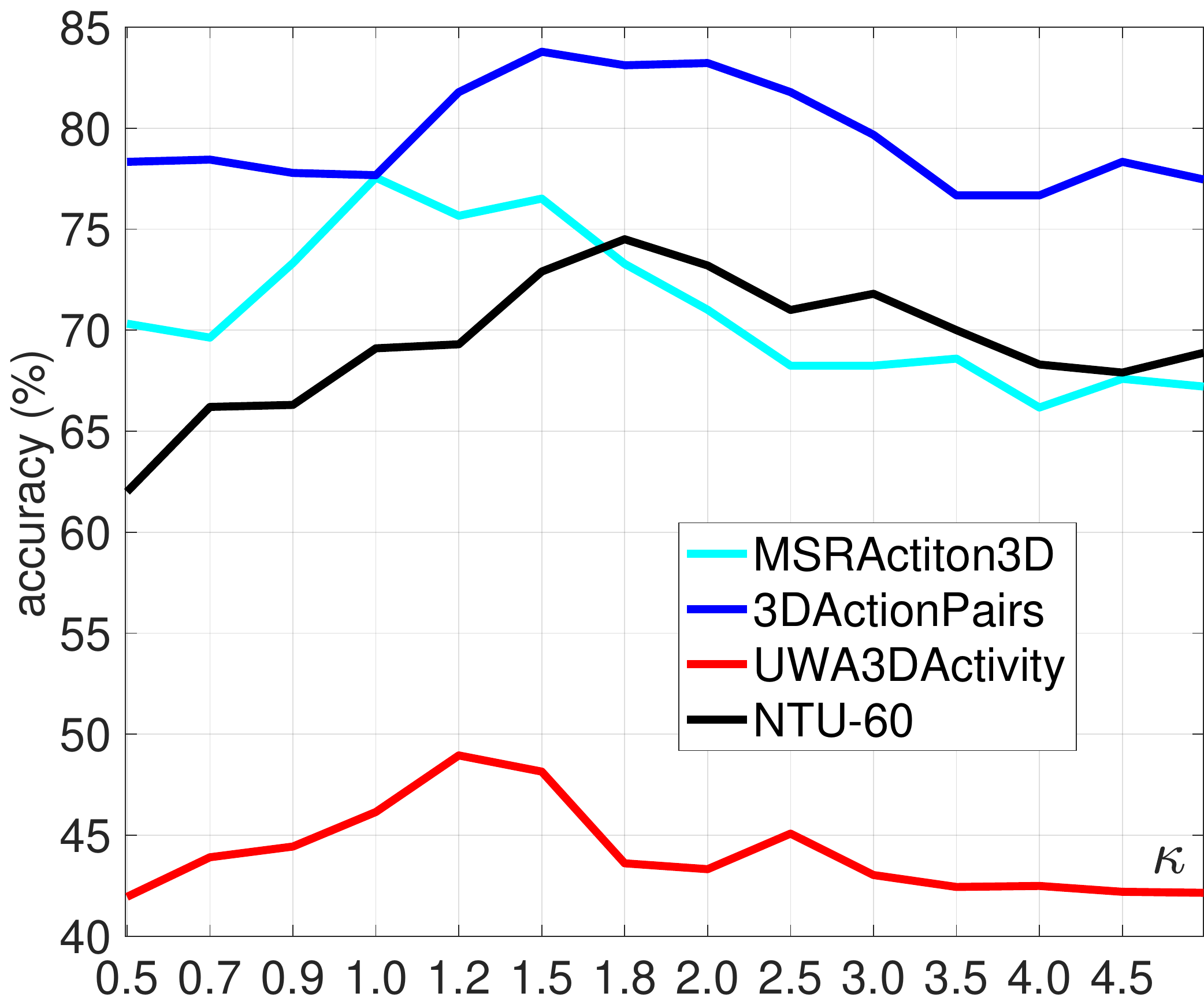}
\caption{\label{fig:a}}
\end{subfigure}
\begin{subfigure}[t]{0.32\linewidth}
\includegraphics[width=\linewidth]{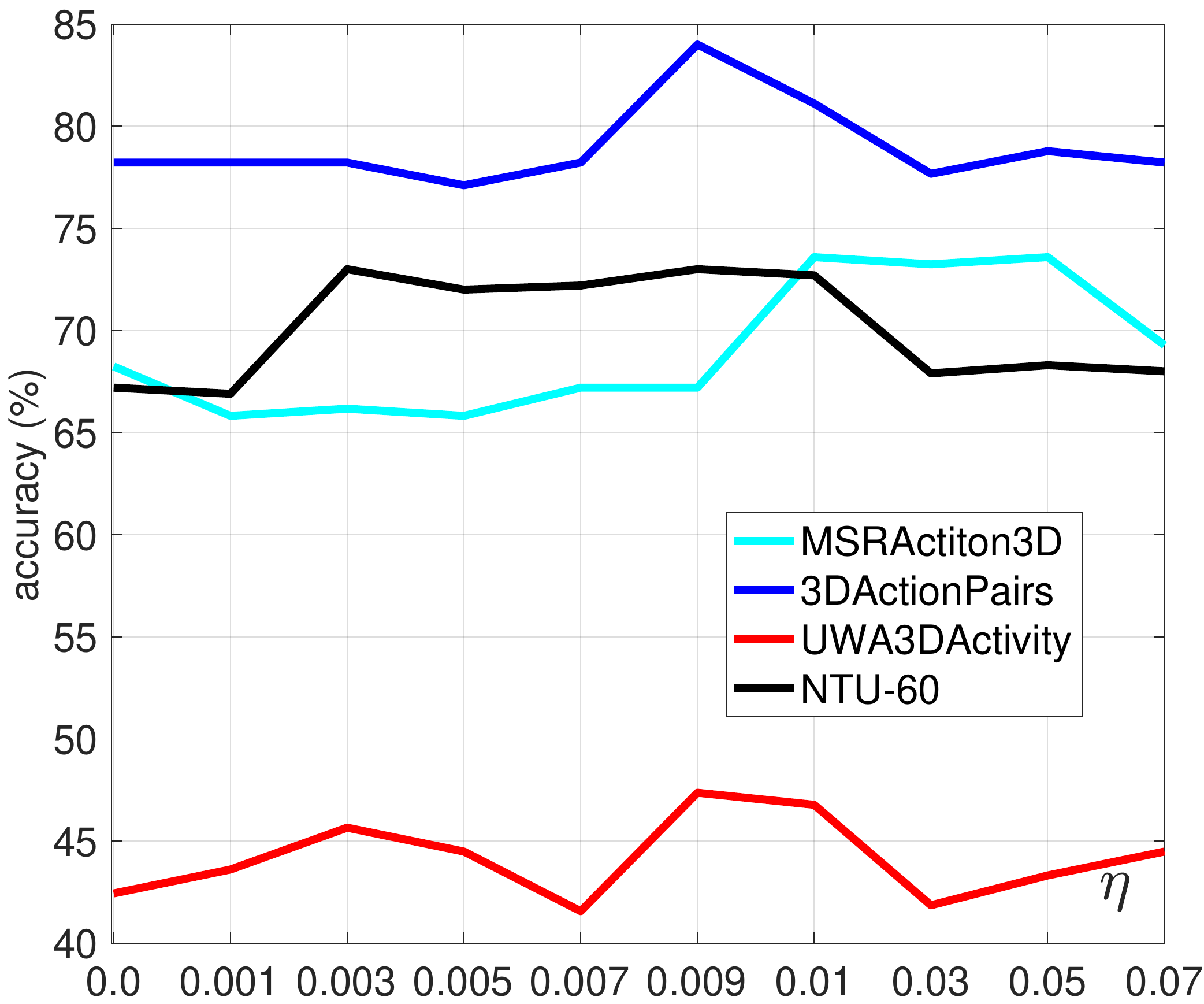}
\caption{\label{fig:b}}
\end{subfigure}
\begin{subfigure}[t]{0.32\linewidth}
\includegraphics[width=\linewidth]{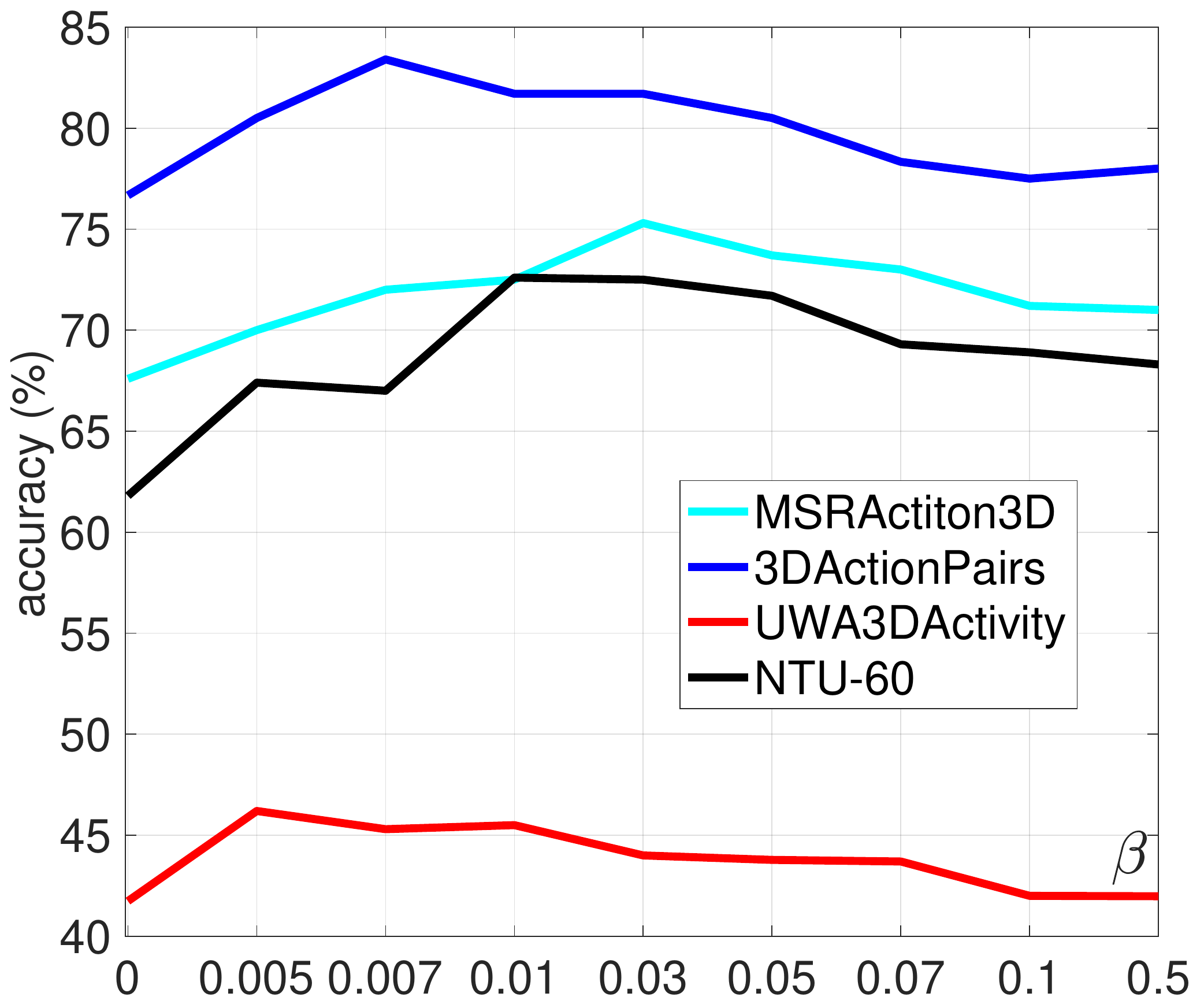}
\caption{\label{fig:beta_eval}}
\end{subfigure}
\caption{Evaluation of (a) $\kappa$ which controls the maximum magnitude and (b) $\eta$ offset from Eq.~\eqref{eq:sigmoid} in SigmaNet and (c) $\beta$ from Eq.~\eqref{eq:sup_ar}. Note that $\beta\!=\!0$ means no regularization term of uDTW in use. We notice that with the regularization term added to the uDTW, the overall performance is improved.}
\label{fig:a_b}
\end{figure}

Figures \ref{fig:a} and \ref{fig:b} show the impact of $\kappa$ and $\eta$ of the scaled sigmoid function in SigmaNet on both small-scale datasets and the large-scle NTU-60 dataset. We notice that $\kappa\!=\!1.5$ performs the best on the three small-scale datasets and $\kappa\!=\!1.8$ works the best on NTU-60. We choose $\kappa\!=\!1.8$ in the experiments for the large-scale datasets. Moreover, $\eta\!\in\![0.003, 0.01]$ works better on NTU-60, and on the small-scale datasets, $\eta\!=\!0.01$ achieves the best performance; thus we choose $\eta\!=\!0.01$ for the experiments.

\subsection{Evaluation of $\beta$}


Figure~\ref{fig:beta_eval} shows the evaluations of $\beta$ for both small-scale datasets and NTU-60. Firstly, note that $\beta\!=\!0$ means lack of the regularization term of uDTW, which immediately causes the performance deterioration. As shown in the figure, $\beta\!=\!0.05$ performs the best on UWA 3D Activity, $\beta\!=\!0.03$ achieves the best performance on MSR Action 3D and $\beta\!=\!0.007$ works the best on 3D Action Pairs dataset. We use the corresponding best $\beta$ values for the smaller datasets. On NTU-60, $\beta\!\in\![0.01, 0.05]$ performs the best compared to other $\beta$ values, thus we choose $\beta\!=\!0.03$ for the experiments on all large-scale datasets. 

\subsection{Evaluation of warping window width}

Table \ref{tab:warp_window} on ECGFiveDays (from UCR) and NTU-60 (50-class, supervised / unsup. settings) shows that uDTW does not break quicker than sDTW (window size is parametrized by $r$). 
%
Very small $r$ may preclude backpropagating through some  paths (of large distance). For such paths `beyond window', learning uncertainty is limited but this is normal. For similar reasons, choosing the right window size is required by other DTW variants too. Also, if $r$ is very large, large uncertainty score may decrease the distance on multitude of paths by downweighting parts of paths (could lead to strange matching) but as the uncertainty is aggregated into the regularization penalty, this penalty prevents uDTW from unreasonable solutions. Lack of regularization penalty ({\em w/o reg.}) affects the most the unsupervised few-shot learning, while supervised loss can still drive SigmaNet to produce meaningful results.

\begin{table}[!htbp]
\caption{Experimental results on ECGFiveDays (from UCR) and NTU-60 (50-class, supervised / unsup. settings) for different warping window widths.}
\label{tab:warp_window}
\begin{center}
\resizebox{0.9\linewidth}{!}{
\setlength{\tabcolsep}{0.5pt}
\begin{tabular}{l l|ccc|ccc|ccc|ccc}
\toprule
\rowcolor{gray!10}
& & \multicolumn{3}{c|}{$\gamma=0.001$} & \multicolumn{3}{c|}{$\gamma=0.01$} & \multicolumn{3}{c|}{$\gamma=0.1$} & \multicolumn{3}{c}{$\gamma=1$}\\
\cline{3-14}
\rowcolor{gray!10}
& & $r$=1.0 & $r$=3.0 & $r$=5.0 & $r$=1.0 & $r$=3.0 & $r$=5.0 & $r$=1.0 & $r$=3.0 & $r$=5.0 & $r$=1.0 & $r$=3.0 & $r$=5.0\\
\midrule
\multirow{3}{*}{$\substack{\text{\fontsize{10}{10}\selectfont ECG}\\\text{\fontsize{10}{10}\selectfont FiveDays}}$
}$\;$& sDTW & 83.4 &  82.8 & 82.0 & 79.7 & 76.8 & 77.8 & 75.4 & 69.0 & 65.3 & 62.5 & 61.7 & 60.2\\
& uDTW & 85.6 & 91.2 & 81.0 & 93.5 & 82.8 & 80.6 & 79.7 & 73.9 & 67.3 & 69.0 & 65.3 & 62.5\\
& $\substack{\text{\fontsize{10}{10}\selectfont uDTW}\\\text{\fontsize{10}{10}\selectfont {\it w/o reg.}}}$ & 75.4 & 74.0 & 69.0 & 79.7 & 77.9 & 76.8 & 65.3 & 62.5 & 61.5 & 61.2 & 62.0 & 60.2\\
\hline
\multirow{3}{*}{$\substack{\text{\fontsize{10}{10}\selectfont NTU-60}\\\text{\fontsize{10}{10}\selectfont (sup.)}}$}& sDTW & 65.7 & 64.7 & 64.8 & 65.2 & 67.8 & 63.9 & 60.0 & 58.9 & 54.3 & 54.0 & 52.2 & 52.3\\
& uDTW & 71.5 & 71.0 & 70.0 & 72.4 &  72.4 & 70.0 & 68.3 & 66.7 & 67.8 & 65.7 & 64.8 & 66.8\\
& $\substack{\text{\fontsize{10}{10}\selectfont uDTW}\\\text{\fontsize{10}{10}\selectfont {\it w/o reg.}}}$ & 66.3 & 65.0 & 65.5 & 66.4 &  68.0 & 65.2 & 62.0 & 59.2 & 55.0 & 52.0 & 52.0 & 51.2\\
\cline{2-14}
\multirow{3}{*}{$\substack{\text{\fontsize{10}{10}\selectfont NTU-60}\\\text{\fontsize{10}{10}\selectfont (unsup.)}}$}& sDTW & 56.7 & 53.2 & 50.0 & 61.7 & 61.7 & 60.0 & 54.4 & 52.5 & 52.1 & 48.3 & 45.2 & 40.9\\
& uDTW & 61.0 & 61.5 & 60.7 & 63.3 & 63.0 & 62.5 & 59.2 & 59.0 & 57.3 & 58.0 & 57.2 & 55.7 \\
& $\substack{\text{\fontsize{10}{10}\selectfont uDTW}\\\text{\fontsize{10}{10}\selectfont {\it w/o reg.}}}$ & 50.1 & 49.3 & 47.0 & 55.3 & 54.0 & 51.3 & 44.1 & 42.0 & 40.7 & 42.3 & 40.1 & 35.6\\
\bottomrule
\end{tabular}}
\end{center}
\end{table}

\bibliographystylelatex{splncs04}
\bibliographylatex{egbib}

\includepdf[width=12.2cm,pages=-,nup=1x2]{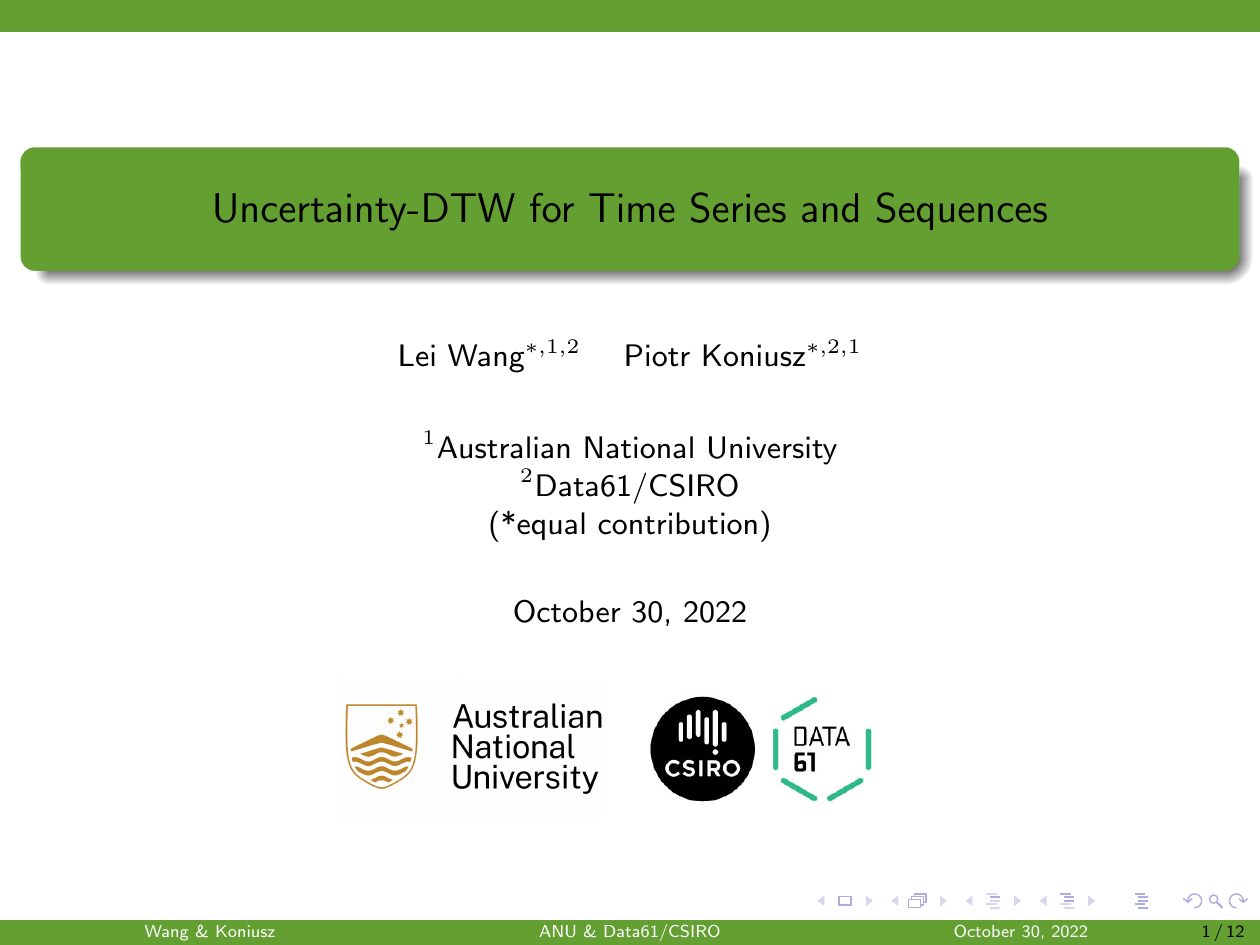}


\end{document}